\documentclass[lettersize,journal]{IEEEtran}
\usepackage{amsmath,amsfonts}
\usepackage{algorithmic}
\usepackage{array}
\usepackage[caption=false,font=normalsize,labelfont=sf,textfont=sf]{subfig}
\usepackage{textcomp}
\usepackage{stfloats}
\usepackage{url}
\usepackage{bigstrut,multirow}
\usepackage{verbatim}
\usepackage{graphicx}
\usepackage{amsmath,amsfonts}
\usepackage{algorithmic}
\usepackage{algorithm}
\usepackage{array}
\usepackage{verbatim}
\usepackage{caption}
\usepackage{import}
\usepackage{colortbl}
\usepackage{booktabs}
\usepackage{xcolor}
\usepackage{orcidlink} 
\usepackage{balance}
\def\BibTeX{{\rm B\kern-.05em{\sc i\kern-.025em b}\kern-.08em
    T\kern-.1667em\lower.7ex\hbox{E}\kern-.125emX}}

\begin{document}
\title{MFCLIP: Multi-modal Fine-grained CLIP for Generalizable Diffusion Face Forgery Detection}
\author{Yaning Zhang\orcidlink{0000-0001-8442-2777}, Tianyi Wang\orcidlink{0000-0003-2920-6099}, \emph{Member, IEEE}, Zitong Yu\orcidlink{0000-0003-0422-6616}, \emph{Senior Member, IEEE}, Zan Gao\orcidlink{0000-0003-2182-5741}, \emph{Senior Member, IEEE}, Linlin Shen\orcidlink{0000-0003-1420-0815}, \emph{Senior Member, IEEE}, and Shengyong Chen\orcidlink{0000-0002-6705-3831}, \emph{Senior Member, IEEE}
\vspace{-2em}
\thanks{This work was supported by the National Natural Science Foundation of China (No.62372325, No.62402255, No.62306061), Shandong Province National Talents Supporting Program (No.2023GJJLJRC-070), Natural Science Foundation of Tianjin Municipality (No.23JCZDJC00280), Shandong Project towards the Integration of Education and Industry (No.801822020100000024), Guangdong Basic and Applied Basic Research Foundation (Grant No. 2023A1515140037), and Open Fund of National Engineering Laboratory for Big Data System Computing Technology (Grant No. SZU-BDSC-OF2024-02). The computational resources are supported by SongShan Lake HPC Center (SSL-HPC) in Great Bay University. (Corresponding author: Zitong Yu and Zan Gao)}

\thanks{Y. Zhang is with Faculty of Computer Science and Technology, Qilu University of Technology (Shandong Academy of Sciences), Jinan, 250014, China. E-mail: zhangyaning0321@163.com}
\thanks{T. Wang is with School of Computing, National University of Singapore, 21 Lower Kent Ridge Rd, 119077, Singapore. E-mail: wangty@nus.edu.sg}
\thanks{Z. Yu is with School of Computing and Information Technology, Great Bay University, Dongguan, 523000, China, also with National Engineering Laboratory for Big Data System Computing Technology, Shenzhen University, Shenzhen 518060, China, and also with Dongguan Key Laboratory for Intelligence and Information Technology. E-mail: yuzitong@gbu.edu.cn}
\thanks{Z. Gao is with Shandong Artificial Intelligence Institute, Qilu University of Technology (Shandong Academy of Sciences), Jinan, 250014, China, and also with the Key Laboratory of Computer Vision and System, Ministry of Education, Tianjin University of Technology, Tianjin, 300384, China. E-mail: zangaonsh4522@gmail.com } 
\thanks{L. Shen is with Computer Vision Institute, College of Computer Science and Software Engineering, Shenzhen University, Shenzhen, 518060, China, also with National Engineering Laboratory for Big Data System Computing Technology, Shenzhen University, Shenzhen, also with Shenzhen Institute of Artificial Intelligence and Robotics for Society, Shenzhen, 518129, China, and also with Guangdong Key Laboratory of Intelligent Information Processing, Shenzhen University, Shenzhen, 518060, China. E-mail: llshen@szu.edu.cn}
\thanks{S. Chen is with Key Laboratory of Computer Vision and System, Ministry of Education, Tianjin University of Technology, Tianjin, 300384, China. E-mail: sy@ieee.org} 
}
\markboth{Journal of \LaTeX\ Class Files,~Vol.~18, No.~9, September~2020}%
{How to Use the IEEEtran \LaTeX \ Templates}

\maketitle

\begin{abstract}
The rapid development of photo-realistic face generation methods has raised significant concerns in society and academia, highlighting the urgent need for robust and generalizable face forgery detection (FFD) techniques. Although existing approaches mainly capture face forgery patterns using image modality, other modalities like fine-grained noises and texts are not fully explored, which limits the generalization capability of the model. In addition, most FFD methods tend to identify facial images generated by GAN, but struggle to detect unseen diffusion-synthesized ones. To address the limitations, we aim to leverage the cutting-edge foundation model, contrastive language-image pre-training (CLIP), to achieve generalizable diffusion face forgery detection (DFFD). In this paper, we propose a novel multi-modal fine-grained CLIP (MFCLIP) model, which mines comprehensive and fine-grained forgery traces across image-noise modalities via language-guided face forgery representation learning, to facilitate the advancement of DFFD. Specifically, we devise a fine-grained language encoder (FLE) that extracts fine global language features from hierarchical text prompts. We design a multi-modal vision encoder (MVE) to capture global image forgery embeddings as well as fine-grained noise forgery patterns extracted from the richest patch, and integrate them to mine general visual forgery traces. Moreover, we build an innovative plug-and-play sample pair attention (SPA) method to emphasize relevant negative pairs and suppress irrelevant ones, allowing cross-modality sample pairs to conduct more flexible alignment. Extensive experiments and visualizations show that our model outperforms the state of the arts on different settings like cross-generator, cross-forgery, and cross-dataset evaluations. Our code will be available at \url{https://github.com/Jenine-321/MFCLIP}.
\end{abstract}


\section{Introduction}
Photorealistic generation models like variational autoencoder (VAE) \cite{vae}, generative adversarial networks (GANs) \cite{NIPS2014_5ca3e9b1, StyleGAN3, StyleGAN2}, denoising diffusion probabilistic models (DDPM) \cite{ddpm}, autoregressive networks \cite{var,yu2022scaling}, and flow-based methods \cite{flowedit} have reached unprecedented progress in synthesizing highly realistic facial images. Thus, advancing face forgery detection (FFD) becomes a critical and urgent demand. 

\begin{figure}[t]
	\centering
	\includegraphics[width=\linewidth]{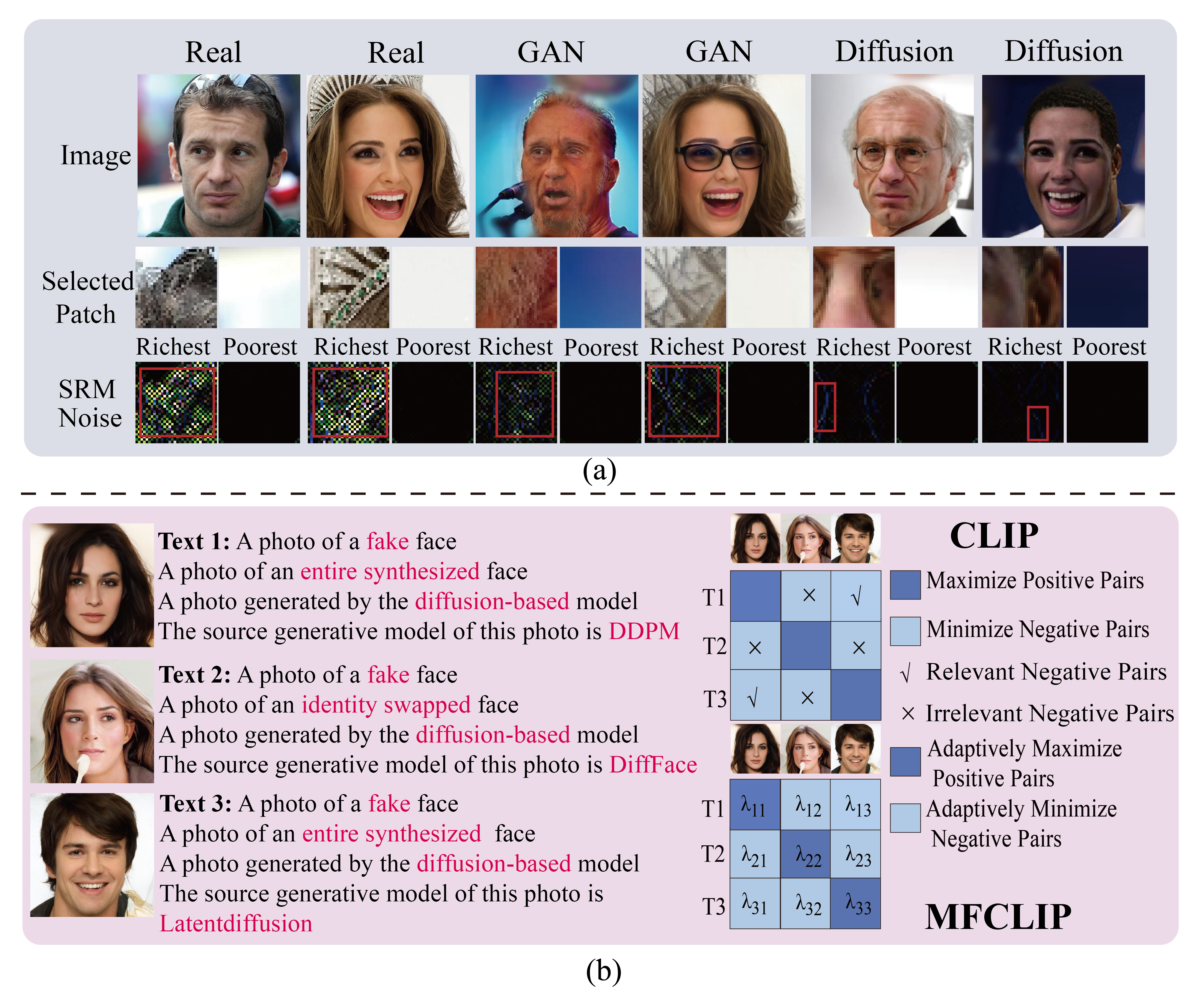} 
	\caption{(a) The visualization of the richest or poorest patch high-frequency noises produced by steganalysis rich model (SRM) \cite{SRM}. The first row represents the RGB images. The second and third rows display the richest or poorest patches and corresponding noises of the real and fake samples of various manipulations, respectively. Inspired by \cite{single}, we split an image into non-overlapping patches, and select the richest patch as well as the poorest patch, respectively. The richest patch is defined as the region with the largest texture diversity, where the texture diversity is measured by the homogeneity of the gray-level co-occurrence matrix (GLCM). (b) Comparison of cross-modal feature alignment between CLIP and MFCLIP.   }\label{fig2} 
\end{figure}

Current FFD methods mainly involve three categories, spatial-based models \cite{Xception,ViT, DIRE,CViT,Sadapter,cpit}, frequency-based models \cite{Freq,gff,SFDG}, and vision-language-based models \cite{VLFFD,DD-VQA}. Spatial-based methods \cite{Xception,CViT} tend to mine domain-specific features in GAN-synthesized images, such as identity and background, which leads to poor generalization to unseen facial images generated by diffusion models. Frequency-based methods \cite{twostream,Freq,noisedf} aim to explore common face forgery traces in the frequency domain, since prior investigations \cite{Frank,Durall} have demonstrated that forgery artifacts produced by various synthesis approaches are mainly concentrated in the high-frequency domain, but they are inclined to capture high-frequency information at a coarse-grained level. As Fig.~\ref{fig2} (a) shows, face images synthesized by advanced diffusion-based models are so realistic that high-frequency noises are rarely observed. Only coarse-grained extraction of high-frequency noises is insufficient to dig comprehensive and universal forgery artifacts. Vision-language-based methods \cite{VLFFD,DD-VQA} intend to study general face manipulation patterns using text-guided image forgery representation learning, but they struggle to introduce hierarchical fine-grained text prompts and achieve flexible cross-modal alignment, to facilitate the generalization to face images synthesized using diffusion models.

\begin{figure*}[t!]
	\centering
	\includegraphics[width=\linewidth]{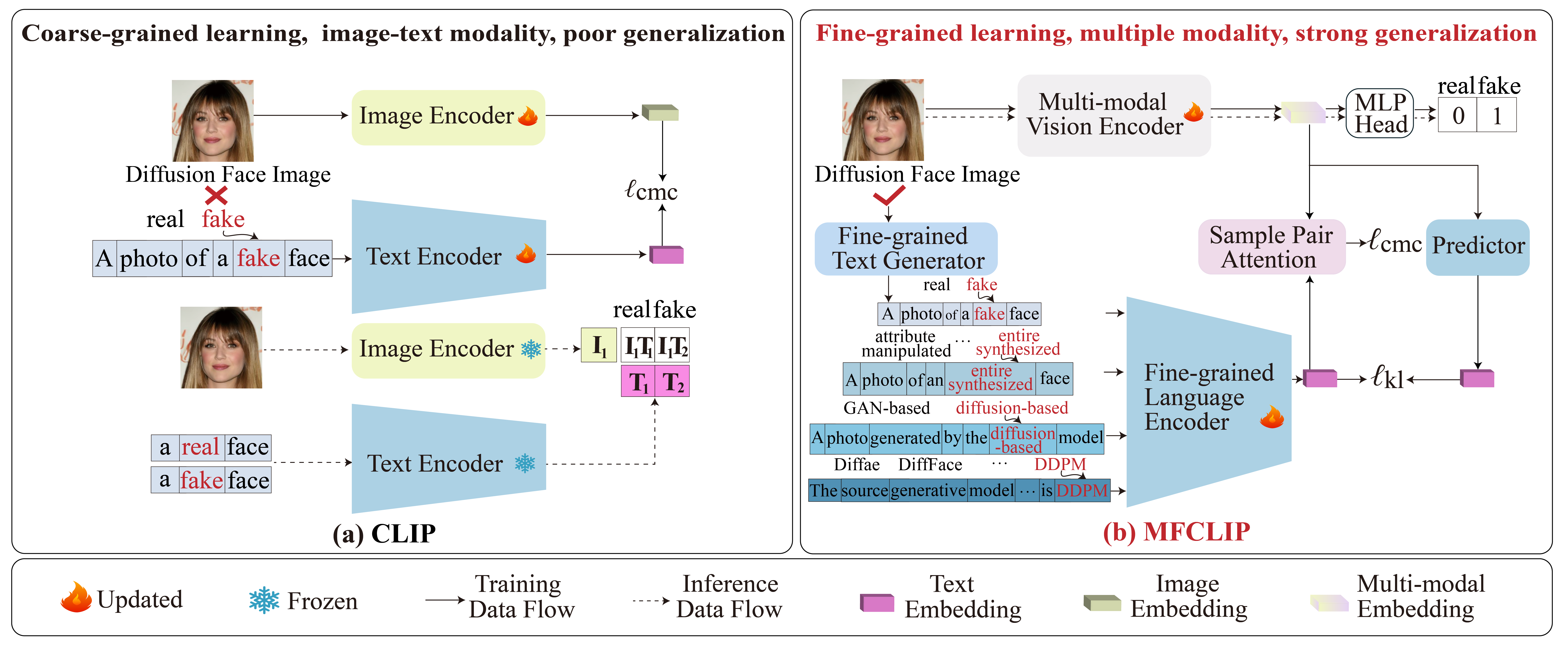} 
	\caption{An overview of the proposed MFCLIP. (a) The vanilla CLIP model tends to extract coarse-grained text embeddings, and only focuses on image-text modality, which leads to poor generalization to diffusion face images. (b) Our MFCLIP model captures fine-grained semantic information, and mines general multi-modal manipulated patterns via text-guided forgery representation learning, to achieve strong generalization to diffusion face images. }\label{fig1}
\end{figure*}
Based on the aforementioned discussion, we propose to introduce fine-grained text prompts to boost the learning of common visual forgery patterns across noise and image modalities via adaptive cross-modal feature alignment, to achieve generalizable diffusion face forgery detection (DFFD). In this paper, inspired by the cutting-edge contrastive language-image pre-training (CLIP) \cite{clip} model, we design a multi-modal fine-grained CLIP (MFCLIP). Unlike CLIP which is pre-trained on large-scale, general natural images, and then fine-tuned to improve inference on downstream tasks like FFD, our MFCLIP model is trained from scratch in an end-to-end manner using facial images generated by various generators like diffusion, to facilitate the advancement of DFFD. Specifically, our MFCLIP model primarily differs from CLIP in the following aspects (see Fig.~\ref{fig1}): First, we observe that there are significant differences between authentic and forgery facial SRM \cite{ SRM} noises from the richest patches (see Fig.~\ref{fig2}), compared to the poorest patches. Specifically, SRM noises extracted from the richest patch are visually evident in real images, but not obvious in fake ones. Besides, the SRM noise in the GAN-generated richest patch is more noticeable than that in the diffusion-generated one. Therefore, we design a multi-modal vision encoder (MVE) with a noise encoder, to study the fine-grained and discriminative noise forgery patterns extracted from the richest patches. Second, we devise a fine-grained text generator (FTG) to build the hierarchical text prompts, and a fine-grained language encoder (FLE) to capture detailed global relations among text prompts. Finally, inspired by the CLIP, we aim to enhance visual forgery representations through feature alignment between cross-modal sample pairs. However, we notice that CLIP tends to maximize the distance between relevant negative pairs (see Fig.~\ref{fig2} (b)), which is regarded as a sub-optimal feature alignment solution since the relevant negative pairs should be closer to each other in the feature space. To address the limitation, we design a plug-and-play sample pair attention (SPA) module to flexibly emphasize relevant negative pairs and suppress irrelevant ones. To sum up, the contributions of our work are as follows:

$\bullet$ We propose a novel MFCLIP model for generalizable DFFD, which incorporates fine-grained noises extracted from the richest patches with global image forgery artifacts, as well as enhances visual features across image-noise modalities via text-guided face forgery representation learning.

$\bullet$ We devise an innovative plug-and-play sample pair attention (SPA) method to adaptively emphasize relevant negative pairs and suppress irrelevant ones, which can be integrated into any vision-language-based models with only a slight growth in computational costs.

$\bullet$ Extensive experiments and visualizations show that our method outperforms the state of the arts on various protocols such as cross-generator evaluation, cross-forgery evaluation, and cross-dataset evaluation. 

The remainder of the paper is organized as follows: Section~\ref{sec2} introduces related work, and Section~\ref{sec3} explains the proposed MFCLIP method. Section~\ref{sec4} describes the experiments, including experimental settings and comparison with the state of the art. Section~\ref{sec5} analyses the results of the ablation study. Section~\ref{sec6} presents the visualizations, and the conclusion and limitations are discussed in Section~\ref{sec7}.

\vspace{-1em}
\section{Related Work}\label{sec2} 
Our research mainly involves two aspects: face forgery detection and vision-language models. In the following subsections, we discuss the two points, separately.
\vspace{-1em}
\subsection{Face Forgery Detection}
Existing efforts have achieved considerable progress in the field of FFD \cite{DeepfakeSurvey2024Wang}. Some methods \cite{DTN,TransDFD} focus on forgery artifacts in the spatial domain. Rossler et al. \cite{FF++} employ an Xception pre-trained on ImageNet, to capture local spatial forgery artifacts. To explore comprehensive relations among image patches, a convolutional vision transformer (CViT) model \cite{CViT} is designed to combine CNN with vision transformer (ViT) \cite{ViT}  for FFD, to mine global forgery traces in the spatial domain. Diffusion reconstruction error (DIRE) \cite{DIRE} utilizes the discrepancy between an input image and corresponding reconstruction for diffusion-generated image detection. In addition to employing spatial forgery patterns for FFD, there are also some frequency-aware approaches \cite{Freq,SFDG}. Hsu et al. \cite{InpaintingDiffusion} propose a diffusion-based image inpainting method that combines stochastic sampling with feature keypoint guidance, to improve small-area fingerprint recognition. SFDG \cite{SFDG} is proposed to mine the relation-aware representations in spatial and frequency domains using dynamic graph learning. FreqNet \cite{Freq} is devised to focus on high-frequency forgery traces across spatial and channel dimensions. To capture diverse forgery patterns across various modalities, TwoStream \cite{gff} is designed to integrate the high-frequency noise features with RGB content for generalized FFD. Recently, VLFFD \cite{VLFFD} is proposed to achieve generalization and interpretability for FFD, where fine-grained text prompts and coarse-grained original data are collaboratively employed to guide the coarse-and-fine co-training network learning. DD-VQA \cite{DD-VQA} introduces a multi-modal transformer, which boosts the learning of face forgery representations using a text and image contrastive learning scheme. By contrast, since there are significant discrepancies between authentic and forgery facial noise images extracted by SRM from the richest patches, we solely extract the SRM noises from the richest patch, and combine them with global spatial forgery patterns to learn more general embeddings. Furthermore, we boost common forged representations across image and noise modalities via fine-grained vision-language matching.

\vspace{-1em}
\subsection{Vision-Language Models}
Different from conventional image-based models that consist of an image feature encoder and a classifier to predict a fixed set of predefined sample categories, the vision-language models \cite{vls,clip} such as contrastive language-image pre-training (CLIP) \cite{clip} simultaneously train an image encoder and a text encoder, to match image and text pairs from training datasets. CLIP facilitates inference on downstream tasks using a zero-shot linear classifier embedded with the class names or descriptions from the target dataset. To circumvent the labour-intensive and inconsistent performance associated with prompt engineering for text encoders, CoOp \cite{ coop} models the context words of a prompt using learnable vectors, and places the class token (i.e., names or descriptions) at the end or middle position. CFPL \cite{cfpl} is designed to study the various semantic prompts conditioned on different visual features for generalizable face anti-spoofing. Unlike traditional CLIP-based methods, our model is capable of capturing visual face forgery embeddings across image-noise modalities, and aligning the cross-modality features using the SPA method, adaptively. 

\begin{figure*}[t!]
	\centering
	\includegraphics[width=\linewidth]{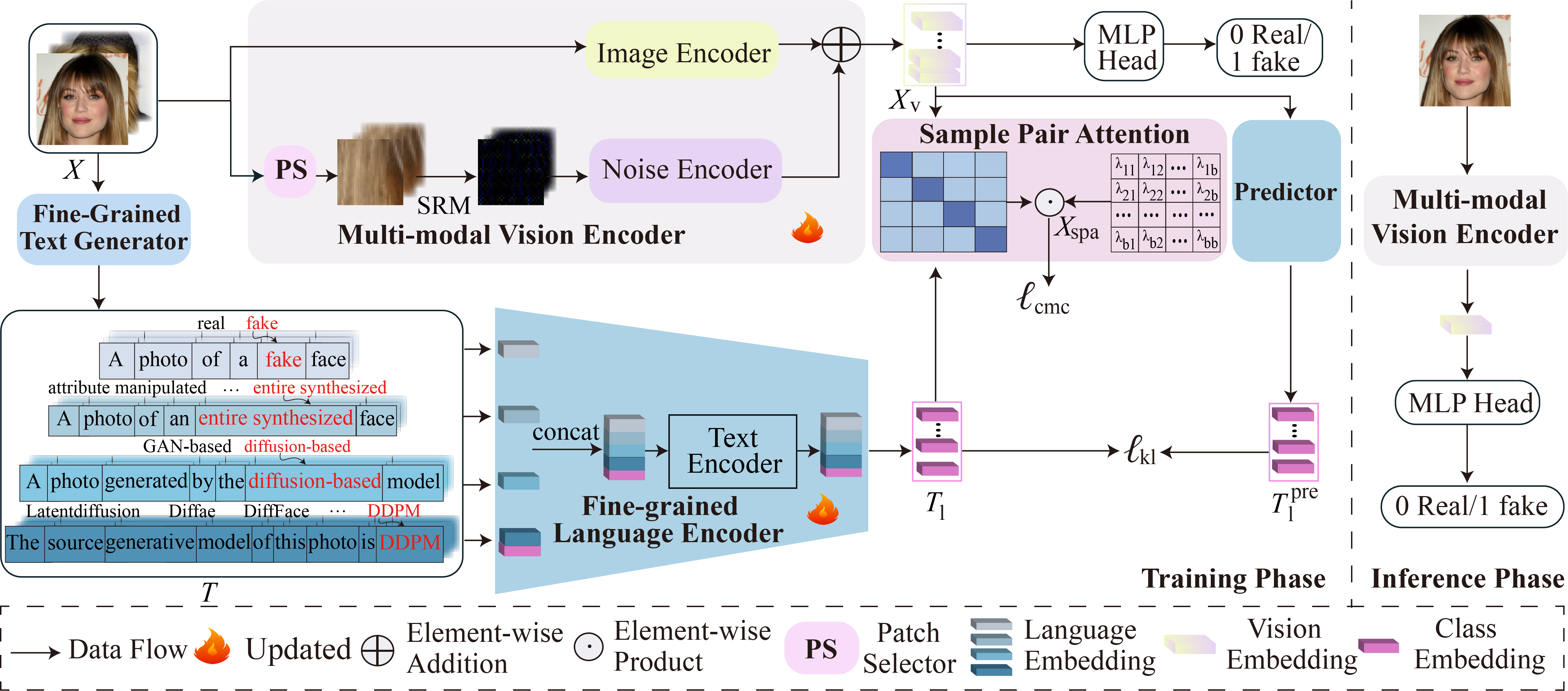} 
	\caption{The framework of MFCLIP. We first generate the hierarchical text prompts through the FTG module, to build the image-text pairs. We then send them to MVE to explore general visual forgery representations across image-noise modalities, and FLE to capture the detailed text embeddings, respectively. Subsequently, they are transferred to a SPA module to align features between cross-modal sample pairs, adaptively. Meanwhile, the visual features are fed into the predictor to predict text embeddings, and the MLP head to acquire the final prediction, respectively. 
	}\label{fig4}
\end{figure*}
\vspace{-2.5em}
\section{Methodology} \label{sec3}

\subsection{Method Overview}

To capture general multi-modal visual forgery patterns for generalizable DFFD, we design the multi-modal fine-grained CLIP model, to conduct the fine-grained text-guided visual forgery representation learning. Different from vision-language-based methods that focus on image and text modalities, our model explores fine-grained noise, image, and text modalities. As Fig.~\ref{fig4} shows, MFCLIP mainly consists of four key modules: fine-grained text generator (FTG), multi-modal vision encoder (MVE), fine-grained language encoder (FLE), and the sample pair attention (SPA) module. During the training phase, given a batch of input facial images $X$, MFCLIP generates the corresponding text prompts $T$ of each image from the GenFace dataset $D$ via the FTG module, and obtains the image-text pairs $(X,T)\in D$ with one-hot labels $y\in\big\{[0,1]^T,[1,0]^T\}$. MFCLIP then captures the multi-modal visual forgery embeddings $X_\text{v}$ through the MVE module, and extracts abundant fine-grained language embeddings $T_\text{l}$ via the FLE module, respectively. Thereafter, $X_\text{v}$ and $T_\text{l}$ are transmitted to the SPA module to adaptively emphasize and suppress sample pairs, to generate $X_\text{spa}$. Meanwhile, $X_\text{v}$ is fed into a predictor with a fully connected layer to predict language features $T_\text{l}^\text{pre}$. Finally, the multilayer perceptron (MLP) head consisting of a full connection layer generates the final predictions $y_\text{pre}$. The kullback-leibler (KL)  divergence loss function $\mathcal{L}_\text{kl}$ is leveraged to measure the difference between predicted language features $T_\text{l}^\text{pre}$ and the true language embeddings $T_\text{l}$. The cross-modality contrastive (CMC) loss $\mathcal{L}_\text{cmc}$ is used to align features between vision-language sample pairs. The cross-entropy (CE) loss $\mathcal{L}_\text{ce}$ is utilized to compute the discrepancy between the predicted label $y_\text{pre}$ and the ground truth label $y$. That is, 
\begin{align}
	\text{MFCLIP}(X,T) &= \text{SPA}(\text{MVE}(X),\text{FLE}(T)) \nonumber \\
	&=\text{SPA}(X_\text{v},T_\text{l})=X_\text{spa} \\
	\mathcal{L}(X,T)\ &=\ \mathcal{L}_\text{kl}\ (T_\text{l}^\text{pre},T_\text{l})+\mathcal{L}_\text{cmc}(X_\text{spa})+\mathcal{L}_\text{ce}(y_\text{pre},y)
\end{align}

During the inference phase, MFCLIP adopts MVE to generate the visual forgery pattern across image-noise modalities, which is then fed into the MLP head to yield the final prediction. It can be expressed formally as follows:
\begin{align}
	\text{MFCLIP}(X)&=\text{MLP}(\text{MVE}(X))\nonumber \\
	&=\text{MLP}(X_{\text{v}}) \nonumber \\
	&=y_\text{pre}   	 	
\end{align}

\subsection{Fine-grained Text Generator}

\begin{figure}[t]
	\centering
	\includegraphics[width=\linewidth]{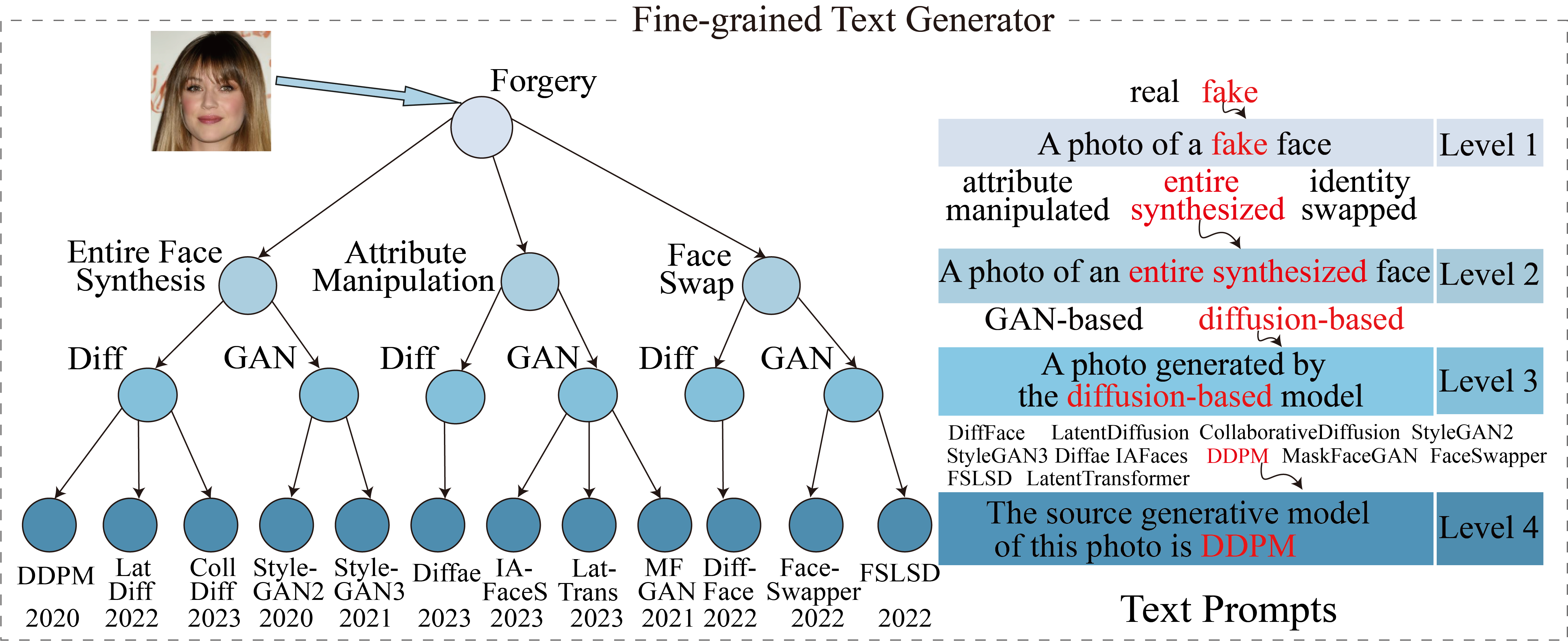} 
	\caption{The schematic diagram of the fine-grained text generator module.  }\label{fig3}
\end{figure}

To create the hierarchical fine-grained text prompts, we design the fine-grained text generator (FTG) module. As Fig.~\ref{fig3} shows, we generate corresponding text prompts for each image based on the hierarchical fine-grained labels provided by the GenFace dataset, to build the image-text pairs. Specifically, we create corresponding text descriptions for each hierarchical level of an image. At level 1, we formulate the texts ``a photo of a real face'' or ``a photo of a fake face". The second level describes the forged images into three types, i.e., ``a photo of an identity swapped face", ``a photo of an attributed manipulated face", and ``a photo of an entire synthesized face". We then generate the text prompts ``a photo generated by the diffusion-based model" or ``a photo generated by the GAN-based model" at level 3. The level 4 refers to the description of the specific generators. We then feed image-text pairs to MVE and FLE, respectively. 

\subsection{Multi-modal Vision Encoder}
Unlike multi-modal FFD models \cite{M2TR,twostream} that mainly concentrate on the interaction between the frequency or noise and RGB information at a coarse-grained level, we design the multi-modal vision encoder (MVE) to capture global image forgery traces as well as fine-grained noise patterns extracted from the richest patches, and integrate them simply and effectively.

As illustrated in Fig.~\ref{fig4}, MVE mainly consists of the image encoder (IE) and noise encoder (NE). IE is composed of a convolutional vision transformer (CViT) \cite{CViT} model, and NE contains a CNN backbone with stacked convolutional layers and a noise transformer encoder (NoT) with $B$ transformer blocks, $\text{TB}_j^\text{n}$, $j=1,2,…, B$. 

\begin{figure}[t!]
	
	\centering
	\includegraphics[width=\linewidth]{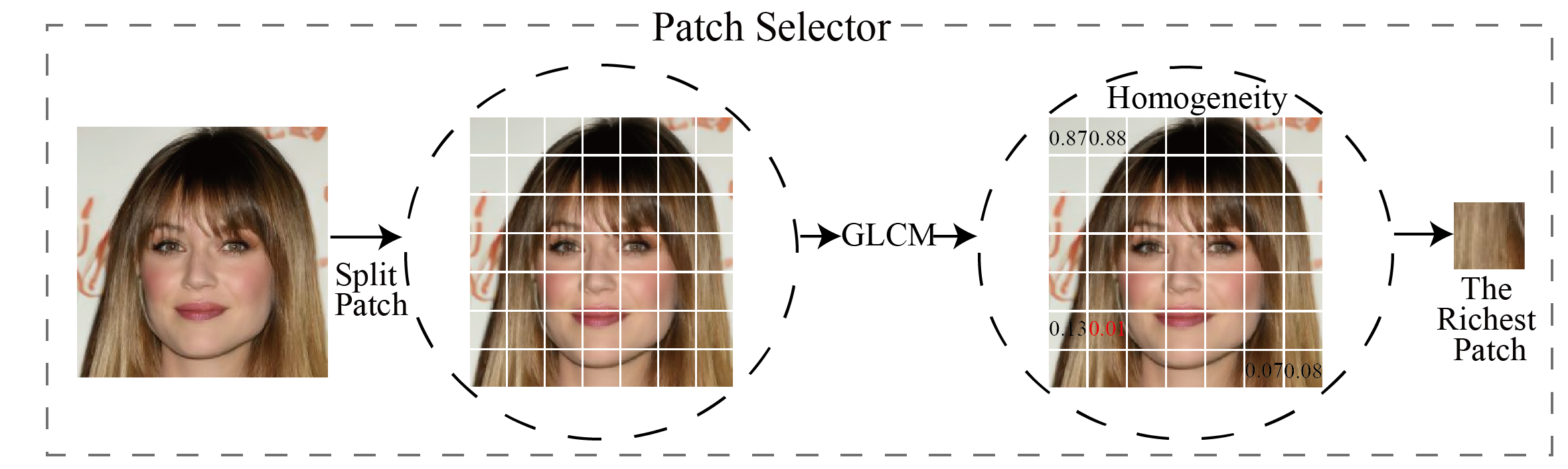} 
	\caption{The pipeline of the patch selector module. The smaller the homogeneity score, the richer the texture diversity.}\label{figrpg}
	
\end{figure}

Specifically, given a batch of $b$ facial images $X\in\mathbb{R}^{b\times3\times224\times224}$, MVE obtains global spatial forgery traces $X_\text{i}\in\mathbb{R}^{b\times d}$ via IE, i.e., $X_\text{i}\ =\ \text{IE}\text(X)$, and the richest patches using the patch selector (PS) module (see Fig.~\ref{figrpg}), where each image  $X_m$, $m=1,...,b$ is divided into non-overlapping $n$ patches $\left\{X_{mi}^\text{p}\in\mathbb{R}^{3\times p\times p}\right\}_{i=1}^n\ $ with the size of $ p\times p$, and the richest patch $X^\text{rp}\in\mathbb{R}^{b\times3\times p\times p}$ is selected. The richest patch refers to the region of the largest texture diversity, where the texture diversity is measured by the homogeneity of GLCM. After that, MVE extracts the noises $N\in\mathbb{R}^{b\times3\times p\times p}$ from the richest patch $X^\text{rp}$ using SRM, and then captures the comprehensive noise forgery patterns through the noise encoder (NE). In detail, given the noises $N$, NE first extracts local noise embeddings $N_\text{loc}\in\mathbb{R}^{b\times c\times h\times w}$with channel $c$, height $h$, and width $w$ from $N$ via the CNN backbone, to conduct feature alignment. $N_\text{loc}$ is then flattened and projected to a 2D token with the dimension of $d$ along the channel. After that, it is appended with a learnable class token to mine the comprehensive noise forgery patterns, to obtain $N_\text{tok}=\text{App}(\text{Proj}(\text{Flat}(\text{Bab}(N))))\in\mathbb{R}^{b\times2\times d}$, and then added with a learnable position embedding$P_\text{n}\in\mathbb{R}^{b\times2\times d}$  to encode the position information. That is,
\begin{align}
	N_1^\text{tra}=N_\text{tok}+P_\text{n}.
\end{align}
Thereafter, it is sequentially fed into $B$ blocks, i.e.,
\begin{align}
	\text{NoT}(N_1^{\text{tra}})
	&=\text{TB}_B^\text{n}\circ\text{TB}_{B-1}^\text{n}\circ \cdots \circ  \text{TB}_2^\text{n} \circ\text{TB}_1^\text{n}(N_1^\text{tra}) \nonumber \\
	&=\text{TB}_B^\text{n}\circ\text{TB}_{B-1}^\text{n}\circ \cdots \circ \text{TB}_2^\text{n}(N_2^\text{tra})
	\nonumber \\
	&= \cdots =\text{TB}_B^\text{n}\circ\text{TB}_{B-1}^\text{n}(N_{B-1}^\text{tra})\nonumber \\
	&= \text{TB}_B^\text{n}(N_B^\text{tra})     \nonumber \\
	&=N_\text{NoT},
\end{align}
where $\circ$ denotes function decomposition.
NE captures the extensive noise forgery traces $N_\text{n}\in\mathbb{R}^{b\times d} $ using the class token in $ N_\text{NoT}$. MVE finally yields the visual forgery features $X_\text{v}$ across image-noise modalities by adding $X_\text{i}$ with $N_\text{n}$ element-wisely, i.e., $X_\text{v}=X_\text{i} + N_\text{n}$, which are then fed into the SPA module, predictor, and the MLP head, respectively.

\subsection{Fine-grained Language Encoder}
To extract hierarchical fine-grained text embeddings, we devise the fine-grained language encoder (FLE), which consists of a text encoder (TE) with $L$ transformer blocks ${\text{TB}_j^\text{t}}$, $j=1,2,…, L$. Specifically, given a batch of text prompts $T$, each of which $T_m$ contains four sentences $\{T_{mi}^\text{sen}\}_{i=1}^4$. For each sentence $T_{mi}^\text{sen}$, we use the tokenizer \cite{clip} to obtain a sequence of word tokens $T_{mi}^\text{tok}\in\mathbb{R}^{77}$ , which is then mapped to the word embedding vector $T_{mi}^\text{emb}\in\mathbb{R}^{77\times d} $ to obtain ${\ T}_i^\text{emb}\in\mathbb{R}^{b\times77\times d}$. We concatenate four sentences to yield ${ T}^\text{emb}\in\mathbb{R}^{b\times308\times d} $, which is then added with position embedding $P_\text{t}\in\mathbb{R}^{b\times308\times d}$, i.e., $T_1^\text{tra}={\ T}^\text{emb}+P_\text{t}$. Thereafter, it is sequentially fed into $L$ blocks, i.e., 
 \begin{align}
	\text{TE}(T_1^{\text{tra}})
	&=\text{TB}_L^\text{t}\circ\text{TB}_{L-1}^\text{t}\circ \cdots \circ  \text{TB}_2^\text{t} \circ\text{TB}_1^\text{t}(T_1^\text{tra}) \nonumber \\
	&=\text{TB}_L^\text{t}\circ\text{TB}_{L-1}^\text{t}\circ \cdots \circ \text{TB}_2^\text{t}(T_2^\text{tra})
	\nonumber \\
	&= \cdots =\text{TB}_L^\text{t}\circ\text{TB}_{L-1}^\text{t}(T_{L-1}^\text{tra})\nonumber \\
	&= \text{TB}_L^\text{t}(T_L^\text{tra})     \nonumber \\
	&=T_\text{fg}.
\end{align}
FLE outputs the fine-grained global language embeddings $T_\text{l}\in\mathbb{R}^{b\times d} $ using the last class embedding in $T_\text{fg}$, which are then transmitted to the SPA module.

\subsection{Sample Pair Attention}
To enhance cross-modal feature alignment flexibly, we propose the sample pair attention (SPA) module. Unlike traditional vision-language-based models like vanilla CLIP \cite{clip}, VLFFD \cite{VLFFD}, and DD-VQA \cite{DD-VQA}, which adjust features using the static alignment mechanism,  MFCLIP dynamically emphasizes relevant sample pairs and suppresses irrelevant ones, enabling more flexible and effective cross-modal representation learning simply and efficiently.

As Fig.~\ref{fig4} illustrates, given a batch of visual forgery features $X_\text{v}\in\mathbb{R}^{b\times d}$ and corresponding fine-grained language representations $T_\text{l}\in\mathbb{R}^{b\times d}$, we obtain a batch of $b $ vision-language sample pairs $\left\{{(X}_\text{v}^i{,T}_\text{l}^i)\in\mathbb{R}^d\right\}_{i=1}^b$. CLIP uses the InfoNCE loss to conduct the cross-modal feature alignment, which pulls the positive pairs together while pushing negative ones apart. However, if some correlated negative pairs are pushed away, noises may be introduced. To address this limitation, MFCLIP generates the vision-language cosine similarity matrix via the SPA method, to emphasize or suppress sample pairs, adaptively. 
\begin{align}
	S_\text{v2l}(X_\text{v}, T_\text{l})= {(X}_\text{v}T_\text{l}^T)\odot\sigma( A)\in\mathbb{R}^{d\times d},
\end{align}
where $A\ \in\mathbb{R}^{d\times d}$ is a learnable weight matrix, $\odot$ denotes a element-wise product, and $\sigma $ is a sigmoid function.
Similarly, MFCLIP generates the language-vision cosine similarity matrix,
\begin{align}
	S_\text{l2v}({T_\text{l},X}_\text{v})= {(T}_\text{l}X_\text{v}^T)\odot\sigma( A )\in\mathbb{R}^{d\times d}.
\end{align}
That is, for the $i$-th pair, the normalized vision-language similarity vector  and the language-vision one are defined as below:
\begin{align}
	S_\text{v2l}^{ij}(X_\text{v}, T_\text{l})=\frac{\text{exp}(\text{sim}{(X}_\text{v}^i{,T}_\text{l}^j)/\tau)}{\sum_{j=1}^{b}{\text{exp}(\text{sim}{(X}_\text{v}^i{,T}_\text{l}^j)/\tau)}}\odot\sigma(A^{ij}), \\
	S_\text{l2v}^{ij}({T_\text{l},X}_\text{v}) = \frac{\text{exp}(\text{sim}{(T}_\text{l}^i{,X}_\text{v}^j)/\tau)}{\sum_{j=1}^{b}{\text{exp}(\text{sim}{(T}_\text{l}^i{,X}_\text{v}^j)/\tau)}}\odot\sigma(A^{ij}),
\end{align}
where $\tau$ is a learnable temperature parameter initialized with 0.07, and the function sim(·) performs a dot product to compute the similarity scores.

\subsection{Loss Function}

{\bfseries\setlength\parindent{0em}  Cross-modal contrastive loss.}
To minimize the distance between positive pairs, while pushing negative pairs away, we introduce the cross-modal contrastive loss function for feature alignment. The one-hot label $y_\text{pa}$ of the $i$-th pair is denoted as $y_\text{pa}^i=\{{y_\text{pa}^{ij}}\}_{j=1}^b$, $y_\text{pa}^{ii}=1$,
$y_\text{pa}^{ij,i\neq j}=0$,
\begin{align}
	\mathcal{L}_\text{v2l}\left(X,T\right)&=\frac{1}{b}\sum_{i=1}^{b}\text{CE}(S_\text{v2l}^i\left(X_\text{v},T_\text{l}\right),y_\text{pa}^i) \nonumber \\
	&=\frac{1}{b}\sum_{i=1}^{b}{-{y_\text{pa}^i}^T\text{log}(}S_\text{v2l}^i\left(X_\text{v},T_\text{l}\right)),\\
	\mathcal{L}_\text{l2v}(T,X)&=\frac{1}{b}\sum_{i=1}^{b}{\text{CE}(S_\text{l2v}^i({T_\text{l},X}_\text{v}),y_\text{pa}^i)} \nonumber \\
	&=\frac{1}{b}\sum_{i=1}^{b}{-{y_\text{pa}^i}^T\text{log}(}S_\text{l2v}^i({T_\text{l},X}_\text{v})),
\end{align}
where CE is cross-entropy. The final cross-modal contrastive loss $\mathcal{L}_\text{cmc}=\ (\mathcal{L}_\text{v2l}+\mathcal{L}_\text{l2v})/2$.

{\bfseries\setlength\parindent{0em}  Kullback-leibler divergence loss.} In order to narrow the gap between the predicted language distribution and the true language distribution, we introduce the kullback-leibler divergence loss. Specifically, we send visual forgery embeddings $X_\text{v}$ to the predictor consisting of a full connection layer, to produce the predicted language representations $T_\text{l}^{\text{pre}}\in\mathbb{R}^{b\times d}$. We leverage the Kullback-leibler divergence loss to bring the predicted language features closer to the authentic ones,
\begin{align}
	\mathcal{L}_\text{kl}= \frac{1}{b}\sum_{i=1}^{b}{\delta{(T_\text{l}^i)}^T\text{log}\frac{{\delta(T}_\text{l}^i)}{\delta(T_\text{l}^{\text{pre}i})}},
\end{align}
where $\delta$ is the softmax function with temperature 0.5, to smooth representations.

{\bfseries\setlength\parindent{0em}  Cross-entropy loss.} To minimize the distance between the predicted label and the true label, we use the cross-entropy loss function. In detail, we transfer $X_\text{v}$ to the MLP head with a full connection layer, to generate the final prediction $y_\text{pre}$. We use the cross-entropy loss as follows:
 \begin{align}
	\mathcal{L}_\text{ce}=\ \frac{1}{b}\sum_{i=1}^{b}{-{y^i}^T\text{log}(}y_\text{pre}^i).
\end{align}

 The total loss is defined as follows:
\begin{align}
	\mathcal{L}=\mathcal{L}_\text{cmc}+\mathcal{L}_\text{ce}+\mathcal{L}_\text{kl}
\end{align}
\begin{table}[t]
	\caption{Cross-forgery generalization. ACC and AUC scores (\%) on remaining manipulations, after training using one manipulation in GenFace. EFS represents entire face synthesis, AM means attribute manipulation, and FS denotes face swapping.  \label{tab1} \vspace{-0.9em}}
	\setlength{\tabcolsep}{0.85mm}{
		\begin{tabular}{cccccccc}     \\\midrule\multirow{3}[6]{*}{Training Set} & \multirow{3}[6]{*}{Model} & \multicolumn{6}{c}{Testing Set } \\\cmidrule{3-8}      &       & \multicolumn{2}{c}{EFS} & \multicolumn{2}{c}{AM} & \multicolumn{2}{c}{FS} \\\cmidrule{3-8}      &       & ACC   & \cellcolor[gray]{0.9}AUC   & ACC   &  \cellcolor[gray]{0.9}AUC   & ACC   &  \cellcolor[gray]{0.9}AUC \\\midrule\multirow{7}[2]{*}{EFS} & Xception \cite{Xception} & -      &     \cellcolor[gray]{0.9}-  & 50.00  & \cellcolor[gray]{0.9}63.14 & 68.06  & \cellcolor[gray]{0.9}79.52 \\      & ViT \cite{ViT}  &      - &   \cellcolor[gray]{0.9} -   & 54.69 & \cellcolor[gray]{0.9}65.86 & 53.13 & \cellcolor[gray]{0.9}61.43  \\      & CViT \cite{CViT}  &      - &  \cellcolor[gray]{0.9}  -   & 50.02 & \cellcolor[gray]{0.9}63.53 & 72.79  & \cellcolor[gray]{0.9}73.82  \\      & DIRE \cite{DIRE}  &  -     &  \cellcolor[gray]{0.9}\cellcolor[gray]{0.9} -    & 50.03  & \cellcolor[gray]{0.9}76.14 & 74.03  & \cellcolor[gray]{0.9}77.89  \\      & FreqNet \cite{Freq}&     -  &     \cellcolor[gray]{0.9}-  & 50.00  & \cellcolor[gray]{0.9}75.41 & 76.48  & \cellcolor[gray]{0.9}69.62  \\      & CLIP \cite{clip} &    -   &  \cellcolor[gray]{0.9}  -   & 56.05  & \cellcolor[gray]{0.9}64.32  & 53.06  & \cellcolor[gray]{0.9}61.40 
			\\      & FatFormer \cite{FatFormer} &    -   &  \cellcolor[gray]{0.9}  -   & 55.89 & \cellcolor[gray]{0.9}66.90  & 56.21  & \cellcolor[gray]{0.9}64.89 
			\\      
			& VLFFD \cite{VLFFD} &    -   &  \cellcolor[gray]{0.9}  -   & 55.31 & \cellcolor[gray]{0.9}73.89  & 69.54  & \cellcolor[gray]{0.9}70.31
			\\      & DD-VQA \cite{DD-VQA} &    -   &  \cellcolor[gray]{0.9}  -   & 55.96 & \cellcolor[gray]{0.9}74.02 & 70.23  & \cellcolor[gray]{0.9}80.02
			\\      & \textbf{MFCLIP (Ours)} &     -  &   \cellcolor[gray]{0.9}  -  & \textbf{58.05}  & \cellcolor[gray]{0.9}\textbf{78.76} & \textbf{76.88 } & \cellcolor[gray]{0.9}\textbf{81.99} \\\midrule\multirow{7}[2]{*}{AM} & Xception \cite{Xception}& 50.20  & \cellcolor[gray]{0.9}51.45  &      - &   \cellcolor[gray]{0.9}-    & 50.11  & \cellcolor[gray]{0.9}54.57  \\      & ViT \cite{ViT}   & 50.29  & \cellcolor[gray]{0.9}60.37  &   -    &  \cellcolor[gray]{0.9}-    & 50.19  & \cellcolor[gray]{0.9}55.04  \\      & CViT \cite{CViT}  & 50.15  & \cellcolor[gray]{0.9}70.32&    -   &   \cellcolor[gray]{0.9}-    & 50.02  & \cellcolor[gray]{0.9}60.74  \\      & DIRE \cite{DIRE} & 51.14  & \cellcolor[gray]{0.9}72.41  &   -    &    \cellcolor[gray]{0.9}-   & 51.24 & \cellcolor[gray]{0.9}70.45  \\      & FreqNet \cite{Freq}& 50.88  & \cellcolor[gray]{0.9}74.68  &   -    &   \cellcolor[gray]{0.9}\cellcolor[gray]{0.9}-    & 50.33  & \cellcolor[gray]{0.9}76.34  \\     & CLIP \cite{clip}  & 52.48  & \cellcolor[gray]{0.9}60.96 &    -   &   \cellcolor[gray]{0.9}-    & 51.01  & \cellcolor[gray]{0.9}55.21  	\\       & FatFormer \cite{FatFormer}  & 51.67 & \cellcolor[gray]{0.9}63.90 &    -   &   \cellcolor[gray]{0.9}-    & 51.82  & \cellcolor[gray]{0.9}59.03	\\   
			& VLFFD \cite{VLFFD} &   52.06   &  \cellcolor[gray]{0.9}  68.98& - & \cellcolor[gray]{0.9}-  & 51.52  & \cellcolor[gray]{0.9}74.40
			\\      & DD-VQA \cite{DD-VQA} &    52.15  &  \cellcolor[gray]{0.9} 73.10  & - & \cellcolor[gray]{0.9}- & 52.07  & \cellcolor[gray]{0.9}77.50
			\\      & \textbf{MFCLIP (Ours)} & \textbf{53.29}&\cellcolor[gray]{0.9}\textbf{ 87.76}&    -   &   \cellcolor[gray]{0.9}-   & \textbf{52.61}  & \cellcolor[gray]{0.9}\textbf{80.31}  \\\midrule\multirow{7}[2]{*}{FS} & Xception \cite{Xception}& 50.42  & \cellcolor[gray]{0.9}76.48  & 53.75  & \cellcolor[gray]{0.9}75.62 &    -   & \cellcolor[gray]{0.9}- \\      & ViT \cite{ViT}     & 51.09  & \cellcolor[gray]{0.9}69.16  & 52.37  & \cellcolor[gray]{0.9}78.11 &   -    & \cellcolor[gray]{0.9}- \\      & CViT \cite{CViT} & 50.22  & \cellcolor[gray]{0.9}73.88  & 49.98 & \cellcolor[gray]{0.9}73.75&   -    & \cellcolor[gray]{0.9}- \\      & DIRE \cite{DIRE}  & 54.06  & \cellcolor[gray]{0.9}79.65  & 52.13  & \cellcolor[gray]{0.9}78.32  &    -   & \cellcolor[gray]{0.9}- \\      & FreqNet \cite{Freq}& 53.46  & \cellcolor[gray]{0.9}73.68  & 51.97  & \cellcolor[gray]{0.9}74.18  &    -   & \cellcolor[gray]{0.9}- \\      & CLIP \cite{clip} & 52.10  & \cellcolor[gray]{0.9}71.47  & 50.16 & \cellcolor[gray]{0.9}62.27 &   -    & \cellcolor[gray]{0.9}- \\     	 & FatFormer \cite{FatFormer} & 54.06  & \cellcolor[gray]{0.9}74.78  & 53.81 & \cellcolor[gray]{0.9}65.89&   -    & \cellcolor[gray]{0.9}- \\     & VLFFD \cite{VLFFD} &   55.38   &  \cellcolor[gray]{0.9}  80.74& 54.60& \cellcolor[gray]{0.9}77.14 & -  & \cellcolor[gray]{0.9}-
			\\      & DD-VQA \cite{DD-VQA} &    56.02  &  \cellcolor[gray]{0.9} 81.10  & 54.79 & \cellcolor[gray]{0.9}76.05& - & \cellcolor[gray]{0.9}-
			\\      &\textbf{MFCLIP (Ours)} & \textbf{60.08} & \cellcolor[gray]{0.9}\textbf{ 84.93} & \textbf{62.58} & \cellcolor[gray]{0.9}\textbf{ 80.38} &     -  & \cellcolor[gray]{0.9}- \\\bottomrule\end{tabular}%
	}
	
	\vspace{-2em}
	
\end{table}

\begin{table}[t!]
	\caption{Cross-generator evaluation on FS. ACC and AUC scores (\%) on remaining generators, after training using one generator in GenFace. FS denotes face swapping. \label{tabfs}}
	\setlength{\tabcolsep}{0.9mm}{
		\begin{tabular}{cccccccc}\toprule\multirow{3}[6]{*}{Training Set} & \multirow{3}[6]{*}{Model} & \multicolumn{6}{c}{Testing Set } \\\cmidrule{3-8}      &       & \multicolumn{2}{c}{DiffFace} & \multicolumn{2}{c}{FSLSD} & \multicolumn{2}{c}{FaceSwapper} \\\cmidrule{3-8}      &       & ACC   & \cellcolor[gray]{0.9}AUC   & ACC   & \cellcolor[gray]{0.9}AUC   & ACC   & \cellcolor[gray]{0.9}AUC \\\midrule\multirow{7}[2]{*}{DiffFace} & Xception \cite{Xception}&     \multicolumn{1}{c}{-}  &   \multicolumn{1}{c}{\cellcolor[gray]{0.9}-}  & 50.00  & \cellcolor[gray]{0.9}48.45 & 50.00  & \cellcolor[gray]{0.9}83.60 \\      & ViT \cite{ViT}  &  \multicolumn{1}{c}{-}     & \multicolumn{1}{c}{\cellcolor[gray]{0.9}-}      & 52.27  & \cellcolor[gray]{0.9}54.47 & 65.58  & \cellcolor[gray]{0.9}86.02  \\      & CViT \cite{CViT} & \multicolumn{1}{c}{-}      &  \multicolumn{1}{c}{\cellcolor[gray]{0.9}-}     & 50.00  & \cellcolor[gray]{0.9}49.28 & 50.04  & \cellcolor[gray]{0.9}79.17  \\      & DIRE \cite{DIRE} &  \multicolumn{1}{c}{-}     & \multicolumn{1}{c}{\cellcolor[gray]{0.9}-}      & 50.00  & \cellcolor[gray]{0.9}55.49 & 50.00  & \cellcolor[gray]{0.9}88.01  \\      & FreqNet \cite{Freq}&  \multicolumn{1}{c}{-}     &  \multicolumn{1}{c}{\cellcolor[gray]{0.9}-}     & 49.75  & \cellcolor[gray]{0.9}44.42 & 49.65  & \cellcolor[gray]{0.9}69.93  \\      & CLIP \cite{clip} &    \multicolumn{1}{c}{-}   & \multicolumn{1}{c}{\cellcolor[gray]{0.9}-}      & 51.06  & \cellcolor[gray]{0.9}55.37 & 70.54  & \cellcolor[gray]{0.9}92.62  \\     
			& FatFormer \cite{FatFormer} &    \multicolumn{1}{c}{-}   & \multicolumn{1}{c}{\cellcolor[gray]{0.9}-}      & 51.95  & \cellcolor[gray]{0.9}57.35 & 71.87  & \cellcolor[gray]{0.9}93.76  \\     
			& VLFFD \cite{VLFFD} &    \multicolumn{1}{c}{-}   & \multicolumn{1}{c}{\cellcolor[gray]{0.9}-}      & 52.44  & \cellcolor[gray]{0.9}56.03 & 72.97  & \cellcolor[gray]{0.9}94.03 \\  
			& DD-VQA \cite{DD-VQA} &    \multicolumn{1}{c}{-}   & \multicolumn{1}{c}{\cellcolor[gray]{0.9}-}      & 52.51  & \cellcolor[gray]{0.9}57.89 & 73.74& \cellcolor[gray]{0.9}95.99  \\     
			& \textbf{MFCLIP (Ours)} &     \multicolumn{1}{c}{-}  &  \multicolumn{1}{c}{\cellcolor[gray]{0.9}-}     & \textbf{55.96}  & \textbf{\cellcolor[gray]{0.9}65.76}  & \textbf{76.52 } & \textbf{\cellcolor[gray]{0.9}99.93} \\\midrule\multirow{7}[2]{*}{FSLSD} & Xception \cite{Xception}& 50.26  & \cellcolor[gray]{0.9}54.22  &      \multicolumn{1}{c}{-} &   \multicolumn{1}{c}{\cellcolor[gray]{0.9}-}    & 51.49  &\cellcolor[gray]{0.9}72.14  \\      & ViT \cite{ViT}  & 50.01  & \cellcolor[gray]{0.9}49.08  &     \multicolumn{1}{c}{-}  & \multicolumn{1}{c}{\cellcolor[gray]{0.9}-}      & 50.21  & \cellcolor[gray]{0.9}64.98  \\      & CViT \cite{CViT}  & 50.03  & \cellcolor[gray]{0.9}47.60  &     \multicolumn{1}{c}{-}  &   \multicolumn{1}{c}{\cellcolor[gray]{0.9}-}    & 50.46  & \cellcolor[gray]{0.9}84.44  \\      & DIRE \cite{DIRE} & 50.00  & \cellcolor[gray]{0.9}51.13  &  \multicolumn{1}{c}{-}     &     \multicolumn{1}{c}{\cellcolor[gray]{0.9}-}  & 50.14  & \cellcolor[gray]{0.9}57.44  \\      & FreqNet \cite{Freq}& 53.44  & \cellcolor[gray]{0.9}55.08  &   \multicolumn{1}{c}{-}    &  \multicolumn{1}{c}{\cellcolor[gray]{0.9}-}     & 49.22  & \cellcolor[gray]{0.9}72.31  \\      & CLIP \cite{clip} & 49.67  & \cellcolor[gray]{0.9}45.77  &     \multicolumn{1}{c}{-}  &     \multicolumn{1}{c}{\cellcolor[gray]{0.9}-}  & 52.37  & \cellcolor[gray]{0.9}72.90  \\  
			& FatFormer \cite{FatFormer} & 52.02 & \cellcolor[gray]{0.9}47.33  &     \multicolumn{1}{c}{-}  &     \multicolumn{1}{c}{\cellcolor[gray]{0.9}-}  & 54.71  & \cellcolor[gray]{0.9}73.09  \\      
			& VLFFD \cite{VLFFD} &    \multicolumn{1}{c}{51.78}   & \multicolumn{1}{c}{\cellcolor[gray]{0.9}53.56}      & -  & \cellcolor[gray]{0.9}- & 53.82  & \cellcolor[gray]{0.9}84.66  \\  
			& DD-VQA \cite{DD-VQA} &    \multicolumn{1}{c}{51.39}   & \multicolumn{1}{c}{\cellcolor[gray]{0.9}54.76}      & -  & \cellcolor[gray]{0.9}-& 54.60  & \cellcolor[gray]{0.9}86.51  \\   
			& \textbf{MFCLIP (Ours)} & \textbf{53.65} & \cellcolor[gray]{0.9}\textbf{55.59}  &   \multicolumn{1}{c}{-}    &     \multicolumn{1}{c}{\cellcolor[gray]{0.9}-}  &\textbf{55.52} &\cellcolor[gray]{0.9}\textbf{92.15 }\\\midrule\multirow{7}[2]{*}{FaceSwapper} & Xception \cite{Xception}& 49.99  & \cellcolor[gray]{0.9}63.86  & 56.60  & \cellcolor[gray]{0.9}45.21 &   \multicolumn{1}{c}{-}    &\multicolumn{1}{c}{\cellcolor[gray]{0.9}-}  \\      & ViT \cite{ViT}    & 49.83  & \cellcolor[gray]{0.9}42.26  & 49.72  & \cellcolor[gray]{0.9}41.21 & \multicolumn{1}{c}{-}      &\multicolumn{1}{c}{\cellcolor[gray]{0.9}-}  \\      & CViT \cite{CViT}   & 50.00  & \cellcolor[gray]{0.9}49.83  & 50.00  & \cellcolor[gray]{0.9}51.97 &  \multicolumn{1}{c}{-}     &\multicolumn{1}{c}{\cellcolor[gray]{0.9}-}  \\      & DIRE \cite{DIRE}  & 50.00  & \cellcolor[gray]{0.9}78.89  & 50.00  & \cellcolor[gray]{0.9}65.41  &  \multicolumn{1}{c}{-}     &\multicolumn{1}{c}{\cellcolor[gray]{0.9}-}  \\      & FreqNet \cite{Freq}& 50.01  & \cellcolor[gray]{0.9}46.86  & 49.82  & \cellcolor[gray]{0.9}46.43  &  \multicolumn{1}{c}{-}     &\multicolumn{1}{c}{\cellcolor[gray]{0.9}-}  \\      & CLIP \cite{clip} & 49.50  & \cellcolor[gray]{0.9}44.68  & 49.01  & \cellcolor[gray]{0.9}45.70  &  \multicolumn{1}{c}{-}     &\multicolumn{1}{c}{\cellcolor[gray]{0.9}-}  \\    
			& FatFormer \cite{FatFormer} & 51.67  & \cellcolor[gray]{0.9}46.83  & 50.71 & \cellcolor[gray]{0.9}47.04  &  \multicolumn{1}{c}{-}     &\multicolumn{1}{c}{\cellcolor[gray]{0.9}-}  \\    
			& VLFFD \cite{VLFFD} &    \multicolumn{1}{c}{50.00}   & \multicolumn{1}{c}{\cellcolor[gray]{0.9}65.82}      & 51.13  & \cellcolor[gray]{0.9}63.48 & -& \cellcolor[gray]{0.9}- \\  
			& DD-VQA \cite{DD-VQA} &    \multicolumn{1}{c}{50.02}   & \multicolumn{1}{c}{\cellcolor[gray]{0.9}65.96}      & 52.80  & \cellcolor[gray]{0.9}59.26 & -  & \cellcolor[gray]{0.9}-  \\   
			& \textbf{MFCLIP (Ours)} & \textbf{52.03}  & \cellcolor[gray]{0.9}\textbf{79.07}  & \textbf{59.08} & \cellcolor[gray]{0.9}\textbf{67.23} &     \multicolumn{1}{c}{-}  & \multicolumn{1}{c}{\cellcolor[gray]{0.9}-} \\\bottomrule\end{tabular}%
	}
		\vspace{-2em}
\end{table}

\vspace{-3em}

\section{Experiments}\label{sec4}
To investigate the performance of our MFCLIP detector, we conduct experiments using five FFD datasets: faceforensics++ (FF++) \cite{FF++}, deepfake detection challenge (DFDC) \cite{DFDC}, Celeb-DF \cite{Celeb-DF}, deeperforensics (DF-1.0) \cite{DF1.0}, and GenFace \cite{genface}. The remainder of this section is organized as follows: 1) the experiment setup including implementation details and datasets is introduced. 2) the performance assessments and comparisons based on five FFD datasets are outlined.

\subsection{Experiment Setup}
\subsubsection{ Implementation details}
We develop the detector using PyTorch on a Tesla V100 GPU with 32GB memory. The time it takes to train MFCLIP is approximately 30 hours, and the model is trained for 30 epochs. The number of blocks $B$ and $L$ in MFCLIP is set to 3 and 6, respectively. The patch size $p$, feature dimension $d$, and batch size $b$ are set to 112, 512, and 24, respectively. We set the channel $c$, height $h$, and width $w$ to 64, 8, and 8, respectively. Our model is trained with the Adam optimizer \cite{Adam} with a learning rate of 1e-4 and weight decay of 1e-3. We use the scheduler to drop the learning rate by ten times every 15 epochs. We use accuracy (ACC) and area under the receiver operating characteristic curve (AUC) as evaluation metrics. We measure fréchet inception distance (FID) \cite{FID} to assess the quality of generated face images, by comparing the feature distributions of real and fake images. The lower FID scores indicate higher similarity and better image quality.
\subsubsection{Datasets}
We use the fine-grained face forgery dataset GenFace \cite{genface} for DFFD, where we conduct the cross-generator and cross-forgery evaluation. We use four deepfake datasets to evaluate the generalization and robustness of networks: FF++ \cite{FF++}, DFDC \cite{DFDC}, Celeb-DF \cite{Celeb-DF}, and DF-1.0 \cite{DF1.0}.

{\bfseries\setlength\parindent{0em}  GenFace.} GenFace contains a large number of facial images generated by different generators, which mainly involve three forgeries: entire face synthesis (EFS), attribute manipulation (AM), and face swapping (FS). EFS proposes to create non-existing forgery faces from random noises using generative methods. AM intends to manipulate the facial attributes of a real image to yield a forgery face. FS refers to swapping the identity of source images with that of target images. Each forgery includes several specific generators. Specifically, EFS comprises diffusion-based generators such as DDPM \cite{ddpm}, LatDiff \cite{lad}, CollDiff \cite{coll}, and GAN-based ones like  StyleGAN2  \cite{StyleGAN2} and StyleGAN3 \cite{StyleGAN3}. AM embraces diffusion-based generators including Diffae \cite{diffae}, as well as GAN-based ones such as LatTrans \cite{Latent}, IAFaces \cite{IA-FaceS}, etc. FS incorporates diffusion-based generators such as DiffFace \cite{difface}, and GAN-based ones like FSLSD \cite{FSLSD} and FaceSwapper \cite{FaceSwapper}. The face image is scaled to a size of 224$\times$224. We conduct the normalization by dividing the image pixel values by 255, without data augmentation operations.

{\bfseries\setlength\parindent{0em}  FF++.} FF++ is a widely used benchmark dataset for FFD, which consists of videos with varying compression levels, including raw, high quality (HQ), and low quality (LQ). We use the HQ videos along with the official splits: 720 videos for training, 140 videos for validation, and 140 videos for testing. We take 20 frames at fixed intervals for each video. We leverage dlib \cite{dlib} to extract square facial regions, and then resize them to a fixed resolution of 224$\times$224 pixels. We conduct the normalization by dividing the image pixel values by 255, without data augmentation operations.

{\bfseries\setlength\parindent{0em}  DFDC.} DFDC is a large-scale dataset containing over 100,000 videos with diverse manipulation techniques and realistic distortions. The data preprocessing way of DFDC is the same as that of FF++.

{\bfseries\setlength\parindent{0em}  Celeb-DF.} Celeb-DF is a high-quality deepfake dataset which contains 590 real videos and 5,639 deepfake ones. The data preprocessing way of Celeb-DF is the same as that of FF++.

{\bfseries\setlength\parindent{0em}  DF-1.0.} DF-1.0 is a large-scale face forgery dataset, comprising 60,000 videos with a total of 17.6 million frames. Various real-world perturbations are used to attain a more challenging benchmark with larger scale and higher diversity. The data preprocessing way of DF-1.0 is the same as that of FF++.

\begin{table*}[t!]
	\caption{Cross-diffusion generalization. ACC and AUC scores (\%) on remaining diffusion-based generators, after training using one diffusion-based generator in GenFace.\label{tabdiff}}
	\setlength{\tabcolsep}{3.0mm}{
		\resizebox{1.0\textwidth}{!}{%
			\begin{tabular}{ccccccccccrr}		
				\toprule
				\multirow{3}[5]{*}{Training Set} & \multirow{3}[5]{*}{Model} & \multicolumn{6}{c}{Testing Set }              &       &       &       &  \\\cmidrule{3-12}      &       & \multicolumn{2}{c}{DDPM} & \multicolumn{2}{c}{LatDiff} & \multicolumn{2}{c}{CollDiff} & \multicolumn{2}{c}{DiffFace} & \multicolumn{2}{c}{Diffae} \\\cmidrule{3-12}      &       & ACC   & \cellcolor[gray]{0.9}AUC   & ACC   &  \cellcolor[gray]{0.9}AUC   & ACC   & \cellcolor[gray]{0.9} AUC   & ACC   &  \cellcolor[gray]{0.9}AUC   & \multicolumn{1}{c}{ACC} & \multicolumn{1}{c}{ \cellcolor[gray]{0.9}AUC} \\\midrule\multirow{7}[1]{*}{DDPM} & Xception \cite{Xception}&     -  &  \cellcolor[gray]{0.9}  -   & 50.00  & \cellcolor[gray]{0.9}63.57 & 50.48  & \cellcolor[gray]{0.9}75.94 & 53.07 & \cellcolor[gray]{0.9}96.80  & \multicolumn{1}{c}{50.07} & \multicolumn{1}{c}{\cellcolor[gray]{0.9}87.49} \\      & ViT \cite{ViT}  &    -   &  \cellcolor[gray]{0.9} -    & 50.67  & \cellcolor[gray]{0.9}54.28 & 52.94  & \cellcolor[gray]{0.9}74.53  & 61.31 & \cellcolor[gray]{0.9}88.58 & \multicolumn{1}{c}{50.21} & \multicolumn{1}{c}{\cellcolor[gray]{0.9}49.17} \\      & CViT \cite{CViT} &   -    &   \cellcolor[gray]{0.9}  -  & 50.00  & \cellcolor[gray]{0.9}74.51 & 50.07  & \cellcolor[gray]{0.9}73.24  & 51.51 & \cellcolor[gray]{0.9}96.21  & \multicolumn{1}{c}{50.28} & \multicolumn{1}{c}{\cellcolor[gray]{0.9}92.62 } \\      & DIRE \cite{DIRE} &  -     & \cellcolor[gray]{0.9}   -   & 50.18  & \cellcolor[gray]{0.9}62.50  & 50.21  & \cellcolor[gray]{0.9}69.95  & 56.54 & \cellcolor[gray]{0.9}94.33  & \multicolumn{1}{c}{51.12} & \multicolumn{1}{c}{\cellcolor[gray]{0.9}88.55 } \\    & CAEL \cite{genface} &  -     & \cellcolor[gray]{0.9}   -   & 52.35 & \cellcolor[gray]{0.9}98.42 &51.10& \cellcolor[gray]{0.9}\textbf{85.29}& 64.57 & \cellcolor[gray]{0.9}99.84 & \multicolumn{1}{c}{68.63} & \multicolumn{1}{c}{\cellcolor[gray]{0.9}97.99} \\      & FreqNet \cite{Freq}&   -    &    \cellcolor[gray]{0.9}-   & 50.00  & \cellcolor[gray]{0.9}49.09 & 50.21  & \cellcolor[gray]{0.9}71.52  & 50.26 & \cellcolor[gray]{0.9}91.94 & \multicolumn{1}{c}{53.77} & \multicolumn{1}{c}{\cellcolor[gray]{0.9}91.76} \\      & CLIP \cite{clip} &    -  &    \cellcolor[gray]{0.9}-   & 51.79  & \cellcolor[gray]{0.9}57.63 & 52.28  & \cellcolor[gray]{0.9}60.95  & 76.71 & \cellcolor[gray]{0.9}93.23 & \multicolumn{1}{c}{49.86} & \multicolumn{1}{c}{\cellcolor[gray]{0.9}45.66} \\  
				& FatFormer \cite{FatFormer} &    -  &    \cellcolor[gray]{0.9}-   & 53.70  & \cellcolor[gray]{0.9}59.43 & 54.66  & \cellcolor[gray]{0.9}63.21  & 78.02 & \cellcolor[gray]{0.9}94.80& \multicolumn{1}{c}{51.65} & \multicolumn{1}{c}{\cellcolor[gray]{0.9}48.27} \\  
				& VLFFD \cite{VLFFD}& -& \cellcolor[gray]{0.9}-  &   51.73    &   \cellcolor[gray]{0.9}72.96   & 53.70& \cellcolor[gray]{0.9}74.91 & 77.46& \cellcolor[gray]{0.9}95.11 & \multicolumn{1}{c}{52.83} & \multicolumn{1}{c}{\cellcolor[gray]{0.9}88.87} \\      & DD-VQA \cite{DD-VQA} & -  & \cellcolor[gray]{0.9}-  &    51.45   &    \cellcolor[gray]{0.9}73.99    & 53.84 & \cellcolor[gray]{0.9}76.08  & 77.93& \cellcolor[gray]{0.9}96.05 & \multicolumn{1}{c}{54.01} & \multicolumn{1}{c}{\cellcolor[gray]{0.9}89.36}\\ 
				& \textbf{MFCLIP (Ours)} &  -     &    \cellcolor[gray]{0.9}-   & \textbf{55.69 } & \cellcolor[gray]{0.9}\textbf{79.80}  & \textbf{ 55.78 } & \cellcolor[gray]{0.9}77.09& \textbf{88.89} & \cellcolor[gray]{0.9}\textbf{99.99} & \multicolumn{1}{c}{\textbf{55.94}} & \multicolumn{1}{c}{\cellcolor[gray]{0.9}\textbf{98.76}}
				\\\midrule
				\multirow{7}[1]{*}{LatDiff} & Xception \cite{Xception}& 50.13  & \cellcolor[gray]{0.9}76.45  &   -    &   \cellcolor[gray]{0.9}-    & 50.00  & \cellcolor[gray]{0.9}37.05  & 51.51 & \cellcolor[gray]{0.9}96.21 & \multicolumn{1}{c}{50.28 } & \multicolumn{1}{c}{\cellcolor[gray]{0.9}92.62} \\      & ViT \cite{ViT}  & 66.60  & \cellcolor[gray]{0.9}80.74  &   -    &   \cellcolor[gray]{0.9}-    & 53.20  & \cellcolor[gray]{0.9}54.09  & 52.93 & \cellcolor[gray]{0.9}57.59 & \multicolumn{1}{c}{57.60 } & \multicolumn{1}{c}{\cellcolor[gray]{0.9}65.33} \\      & CViT \cite{CViT}  & 51.73  & \cellcolor[gray]{0.9}90.76  &  -     &   \cellcolor[gray]{0.9}-    & 50.00  & \cellcolor[gray]{0.9}42.36  & 50.01 & \cellcolor[gray]{0.9}44.50  & \multicolumn{1}{c}{47.88} & \multicolumn{1}{c}{\cellcolor[gray]{0.9}44.04} \\      & DIRE \cite{DIRE}& 50.01  & \cellcolor[gray]{0.9}46.29  &      - &    \cellcolor[gray]{0.9}-   & 50.02  & \cellcolor[gray]{0.9}43.13  & 50.01  & \cellcolor[gray]{0.9}48.51 & \multicolumn{1}{c}{50.05} & \multicolumn{1}{c}{\cellcolor[gray]{0.9}45.98} \\  & CAEL \cite{genface} & 97.66    & \cellcolor[gray]{0.9}99.98   &- & \cellcolor[gray]{0.9}-&50.02 & \cellcolor[gray]{0.9}73.22& 75.00 & \cellcolor[gray]{0.9}99.97  & \multicolumn{1}{c}{67.58} & \multicolumn{1}{c}{\cellcolor[gray]{0.9}99.81} \\  & FreqNet \cite{Freq}& 78.50  & \cellcolor[gray]{0.9}83.32  &   -    &   \cellcolor[gray]{0.9}-    & \textbf{79.04}  & \cellcolor[gray]{0.9}\textbf{89.58}  & 50.18 & \cellcolor[gray]{0.9}89.93 & \multicolumn{1}{c}{38.16} & \multicolumn{1}{c}{\cellcolor[gray]{0.9}37.87} \\      & CLIP \cite{clip} & 62.33  & \cellcolor[gray]{0.9}74.53  &    -   &    \cellcolor[gray]{0.9}-   & 52.99  & \cellcolor[gray]{0.9}53.23  & 55.71 & \cellcolor[gray]{0.9}61.05 & \multicolumn{1}{c}{56.64} & \multicolumn{1}{c}{\cellcolor[gray]{0.9}64.38} \\    
				& FatFormer \cite{FatFormer} & 63.99 & \cellcolor[gray]{0.9}76.26  &    -   &    \cellcolor[gray]{0.9}-   & 54.90  & \cellcolor[gray]{0.9}56.14 & 57.89 & \cellcolor[gray]{0.9}63.04 & \multicolumn{1}{c}{58.92} & \multicolumn{1}{c}{\cellcolor[gray]{0.9}66.03} \\    
				& VLFFD \cite{VLFFD}& 87.35  & \cellcolor[gray]{0.9}92.64  &   -    &   \cellcolor[gray]{0.9}-    & 56.32 & \cellcolor[gray]{0.9}67.81 & 87.24 & \cellcolor[gray]{0.9}89.25 & \multicolumn{1}{c}{68.53} & \multicolumn{1}{c}{\cellcolor[gray]{0.9}93.02} \\      & DD-VQA \cite{DD-VQA} & 88.46  & \cellcolor[gray]{0.9}94.00  &    -   &    \cellcolor[gray]{0.9}-   & 57.84  & \cellcolor[gray]{0.9}73.02  & 88.61 & \cellcolor[gray]{0.9}97.05 & \multicolumn{1}{c}{70.56} & \multicolumn{1}{c}{\cellcolor[gray]{0.9}93.70}\\ 
				& \textbf{MFCLIP (Ours)} & \textbf{99.99} & \cellcolor[gray]{0.9}\textbf{99.99} &      - &   \cellcolor[gray]{0.9}-    & 65.08  & \cellcolor[gray]{0.9}77.07  & \textbf{99.98} & \cellcolor[gray]{0.9}\textbf{99.98} & \multicolumn{1}{c}{\textbf{97.92}} & \multicolumn{1}{c}{\cellcolor[gray]{0.9}\textbf{99.99}} \\\midrule\multirow{7}[2]{*}{CollDiff} & Xception \cite{Xception}& 55.69  & \cellcolor[gray]{0.9}96.31  & 49.98  & \cellcolor[gray]{0.9}70.45 &     -  &  \cellcolor[gray]{0.9}  -   & 50.17 & \cellcolor[gray]{0.9}71.98 & \multicolumn{1}{c}{50.19} & \multicolumn{1}{c}{\cellcolor[gray]{0.9}59.50 } \\      & ViT \cite{ViT}   & 55.06  & \cellcolor[gray]{0.9}61.83  & 51.03  & \cellcolor[gray]{0.9}48.64 &   -    &   \cellcolor[gray]{0.9} -   & 50.26 & \cellcolor[gray]{0.9}50.32 & \multicolumn{1}{c}{50.51} & \multicolumn{1}{c}{\cellcolor[gray]{0.9}49.97} \\      & CViT \cite{CViT}  & 99.74  & \cellcolor[gray]{0.9}99.97  & 49.98  & \cellcolor[gray]{0.9}47.59 &   -    &  \cellcolor[gray]{0.9} -    & 81.97 & \cellcolor[gray]{0.9}98.80  & \multicolumn{1}{c}{90.56} & \multicolumn{1}{c}{\cellcolor[gray]{0.9}99.83} \\      & DIRE \cite{DIRE} & 91.52  & \cellcolor[gray]{0.9}81.05  & 50.02  & \cellcolor[gray]{0.9}65.99  &    -   &     \cellcolor[gray]{0.9}-  & 60.79 & \cellcolor[gray]{0.9}93.68 & \multicolumn{1}{c}{56.85} & \multicolumn{1}{c}{\cellcolor[gray]{0.9}96.86} \\  & CAEL \cite{genface} & 50.19    & \cellcolor[gray]{0.9}76.05   & 50.04 & \cellcolor[gray]{0.9}35.11 &- & \cellcolor[gray]{0.9} -& 55.07 & \cellcolor[gray]{0.9}94.83 & \multicolumn{1}{c}{49.08} & \multicolumn{1}{c}{\cellcolor[gray]{0.9}23.79} \\     & FreqNet \cite{Freq}& 93.48  & \cellcolor[gray]{0.9}85.42  & 49.91  & \cellcolor[gray]{0.9}55.59  &   -    &   \cellcolor[gray]{0.9} -   & 50.04 & \cellcolor[gray]{0.9}61.18 & \multicolumn{1}{c}{\textbf{99.95}} & \multicolumn{1}{c}{\cellcolor[gray]{0.9}99.98} \\      & CLIP \cite{clip} & 51.58  & \cellcolor[gray]{0.9}53.92  & 50.21  & \cellcolor[gray]{0.9}46.17 &     -  &    \cellcolor[gray]{0.9}-   & 50.26 & \cellcolor[gray]{0.9}48.64 & \multicolumn{1}{c}{49.82} & \multicolumn{1}{c}{\cellcolor[gray]{0.9}48.36} \\      
				& FatFormer \cite{FatFormer} & 53.81  & \cellcolor[gray]{0.9}55.52  & 53.90 & \cellcolor[gray]{0.9}49.03 &     -  &    \cellcolor[gray]{0.9}-   & 54.80 & \cellcolor[gray]{0.9}52.71 & \multicolumn{1}{c}{52.90} & \multicolumn{1}{c}{\cellcolor[gray]{0.9}50.65} \\      
				& VLFFD \cite{VLFFD}& 94.25  & \cellcolor[gray]{0.9}96.01  &  56.89   &   \cellcolor[gray]{0.9}64.06   & -  & \cellcolor[gray]{0.9}- & 82.29 & \cellcolor[gray]{0.9}94.67 & \multicolumn{1}{c}{72.43} & \multicolumn{1}{c}{\cellcolor[gray]{0.9}92.99} \\      & DD-VQA \cite{DD-VQA} & 93.20  & \cellcolor[gray]{0.9}97.44  &   55.60   &    \cellcolor[gray]{0.9}70.76   & -  & \cellcolor[gray]{0.9}-  & 83.77 & \cellcolor[gray]{0.9}91.59& \multicolumn{1}{c}{73.08} & \multicolumn{1}{c}{\cellcolor[gray]{0.9}92.94}\\ 
				& \textbf{MFCLIP (Ours)} & \textbf{100}& \cellcolor[gray]{0.9}\textbf{\textbf{100}} & \textbf{99.19} & \cellcolor[gray]{0.9}\textbf{99.97} &     -  &   \cellcolor[gray]{0.9} -   & \textbf{99.63} & \cellcolor[gray]{0.9}\textbf{99.96} & \multicolumn{1}{c}{93.94} & \multicolumn{1}{c}{\cellcolor[gray]{0.9}\textbf{99.99}} \\\midrule\multirow{7}[2]{*}{DiffFace} & Xception \cite{Xception}& 99.98  & \cellcolor[gray]{0.9}99.98  & 50.00  & \cellcolor[gray]{0.9}48.31 & 50.00  & \cellcolor[gray]{0.9}52.97  &   -    &    \cellcolor[gray]{0.9}-   & \multicolumn{1}{c}{50.07} & \multicolumn{1}{c}{\cellcolor[gray]{0.9}87.49} \\      & ViT \cite{ViT}    & 87.00  & \cellcolor[gray]{0.9}97.01  & 49.11  & \cellcolor[gray]{0.9}42.97 & 49.82  & \cellcolor[gray]{0.9}49.00  &    -   &  \cellcolor[gray]{0.9}  -   & \multicolumn{1}{c}{48.74} & \multicolumn{1}{c}{\cellcolor[gray]{0.9}40.90 } \\      & CViT \cite{CViT}  & 50.08  & \cellcolor[gray]{0.9}77.27  & 49.98  & \cellcolor[gray]{0.9}54.22 & 50.04  & \cellcolor[gray]{0.9}59.65  &     -  &\cellcolor[gray]{0.9}  -     & \multicolumn{1}{c}{47.52} & \multicolumn{1}{c}{\cellcolor[gray]{0.9}40.44} \\      & DIRE \cite{DIRE} & 50.04  & \cellcolor[gray]{0.9}75.32  & 50.00  & \cellcolor[gray]{0.9}47.21  & \textbf{59.79}  & \cellcolor[gray]{0.9}\textbf{96.91}  &    -   &   \cellcolor[gray]{0.9} -   & \multicolumn{1}{c}{\textbf{73.55}} & \multicolumn{1}{c}{\cellcolor[gray]{0.9}96.53} \\  & CAEL \cite{genface} &  52.42    & \cellcolor[gray]{0.9}96.90   & 49.95 & \cellcolor[gray]{0.9}73.58 &50.32 & \cellcolor[gray]{0.9}76.82& - & \cellcolor[gray]{0.9}-  & \multicolumn{1}{c}{50.55} & \multicolumn{1}{c}{\cellcolor[gray]{0.9}98.01} \\     & FreqNet \cite{Freq}& 49.85  & \cellcolor[gray]{0.9}82.21  & 51.77  & \cellcolor[gray]{0.9}73.37  & 49.70  & \cellcolor[gray]{0.9}75.62  &      - &   \cellcolor[gray]{0.9} -   & \multicolumn{1}{c}{43.43} & \multicolumn{1}{c}{\cellcolor[gray]{0.9}58.69} \\      & CLIP \cite{clip} & 76.34  & \cellcolor[gray]{0.9}91.74  & 49.98  & \cellcolor[gray]{0.9}50.59 & 51.56  & \cellcolor[gray]{0.9}57.34  &    -   &   \cellcolor[gray]{0.9} -   & \multicolumn{1}{c}{55.90 } & \multicolumn{1}{c}{\cellcolor[gray]{0.9}51.90 } \\     
				& FatFormer \cite{FatFormer} & 79.54  & \cellcolor[gray]{0.9}94.20  & 52.89  & \cellcolor[gray]{0.9}53.94 & 53.76  & \cellcolor[gray]{0.9}59.02  &    -   &   \cellcolor[gray]{0.9} -   & \multicolumn{1}{c}{57.89 } & \multicolumn{1}{c}{\cellcolor[gray]{0.9}54.64} \\     
				& VLFFD \cite{VLFFD}& 77.84 & \cellcolor[gray]{0.9}94.76  &   62.33    &   \cellcolor[gray]{0.9}84.97   & 53.66  & \cellcolor[gray]{0.9}60.23  & - & \cellcolor[gray]{0.9}- & \multicolumn{1}{c}{58.08} & \multicolumn{1}{c}{\cellcolor[gray]{0.9}87.96} \\      & DD-VQA \cite{DD-VQA} & 90.22  & \cellcolor[gray]{0.9}99.98 &    63.75  &    \cellcolor[gray]{0.9}83.88  & 54.07 & \cellcolor[gray]{0.9}59.01 & - & \cellcolor[gray]{0.9}- & \multicolumn{1}{c}{59.89} & \multicolumn{1}{c}{\cellcolor[gray]{0.9}88.40}\\ 
				& \textbf{MFCLIP (Ours)} & \textbf{99.99} & \cellcolor[gray]{0.9}\textbf{99.99} & \textbf{85.32} & \cellcolor[gray]{0.9}\textbf{99.94} & 50.57  & \cellcolor[gray]{0.9}75.40  &    -   &   \cellcolor[gray]{0.9} -    & \multicolumn{1}{c}{52.12} & \multicolumn{1}{c}{\cellcolor[gray]{0.9}\textbf{99.92}} \\\midrule\multirow{7}[2]{*}{Diffae} & Xception \cite{Xception}& 53.96  & \cellcolor[gray]{0.9}94.01  & 49.98  & \cellcolor[gray]{0.9}61.27 & 50.12  & \cellcolor[gray]{0.9}68.04  & 49.99 & \cellcolor[gray]{0.9}52.69 &\multicolumn{1}{c}{-}   &\multicolumn{1}{c}{\cellcolor[gray]{0.9}-} \\      & ViT  \cite{ViT}   & 50.52  & \cellcolor[gray]{0.9}50.04  & 49.45  & \cellcolor[gray]{0.9}45.06 & 49.70  & \cellcolor[gray]{0.9}47.99  & 49.51 & \cellcolor[gray]{0.9}46.50  &\multicolumn{1}{c}{-}     &\multicolumn{1}{c}{\cellcolor[gray]{0.9}-}  \\      & CViT \cite{CViT} & 57.78  & \cellcolor[gray]{0.9}98.04  & 50.23  & \cellcolor[gray]{0.9}83.37 & 50.00  & \cellcolor[gray]{0.9}54.42  & 50.13 & \cellcolor[gray]{0.9}80.12 &  \multicolumn{1}{c}{-}      & \multicolumn{1}{c}{\cellcolor[gray]{0.9}-} \\      & DIRE \cite{DIRE}  & 57.45  & \cellcolor[gray]{0.9}94.21  & 50.07  & \cellcolor[gray]{0.9}62.30  & 50.12  & \cellcolor[gray]{0.9}74.84  & 64.14 & \cellcolor[gray]{0.9}99.02 &  \multicolumn{1}{c}{-}   &  \multicolumn{1}{c}{\cellcolor[gray]{0.9}-}\\  & CAEL \cite{genface} & 62.35    & \cellcolor[gray]{0.9}   99.75  &51.95 & \cellcolor[gray]{0.9}95.74 &50.12 & \cellcolor[gray]{0.9}75.34&50.71 & \cellcolor[gray]{0.9}97.96  & \multicolumn{1}{c}{-} & \multicolumn{1}{c}{\cellcolor[gray]{0.9}-} \\     & FreqNet \cite{Freq}& 53.48  &\cellcolor[gray]{0.9}94.58  & 49.91  & \cellcolor[gray]{0.9}44.41  & 59.86  & \cellcolor[gray]{0.9}53.10  & 50.04 & \cellcolor[gray]{0.9}48.82 &\multicolumn{1}{c}{-}&   \multicolumn{1}{c}{\cellcolor[gray]{0.9}-} \\      & CLIP \cite{clip}  & 50.52  & \cellcolor[gray]{0.9}51.58  & 49.91  &  \cellcolor[gray]{0.9}50.11 & 50.80  & \cellcolor[gray]{0.9}52.07  & 49.97 & \cellcolor[gray]{0.9}50.32 & \multicolumn{1}{c}{-}& \multicolumn{1}{c}{\cellcolor[gray]{0.9}-} \\    
				& FatFormer \cite{FatFormer}  & 53.56 & \cellcolor[gray]{0.9}54.85  & 52.01 &  \cellcolor[gray]{0.9}53.89 & 52.58  & \cellcolor[gray]{0.9}54.05  & 52.57 & \cellcolor[gray]{0.9}53.65 & \multicolumn{1}{c}{-}& \multicolumn{1}{c}{\cellcolor[gray]{0.9}-} \\    
				& VLFFD \cite{VLFFD}& 62.79  & \cellcolor[gray]{0.9}93.43  &   60.77   &   \cellcolor[gray]{0.9}82.54  & 52.78  & \cellcolor[gray]{0.9}65.00  & 63.29 & \cellcolor[gray]{0.9}98.36& \multicolumn{1}{c}{-} & \multicolumn{1}{c}{\cellcolor[gray]{0.9}-} \\      & DD-VQA \cite{DD-VQA} & 64.45 & \cellcolor[gray]{0.9}95.06  &    61.78  &    \cellcolor[gray]{0.9}83.11 & 52.87  & \cellcolor[gray]{0.9}69.57  & 60.99 & \cellcolor[gray]{0.9}96.47 & \multicolumn{1}{c}{-} & \multicolumn{1}{c}{\cellcolor[gray]{0.9}-}\\ 
				& \textbf{MFCLIP (Ours)} & \textbf{98.99} & \cellcolor[gray]{0.9}\textbf{99.99} & \textbf{99.82 }&  \cellcolor[gray]{0.9}\textbf{99.98}  &  \textbf{60.07}   &  \cellcolor[gray]{0.9}\textbf{75.80}   & \textbf{99.99} &  \cellcolor[gray]{0.9}\textbf{99.99} &\multicolumn{1}{c}{-}& \multicolumn{1}{c}{\cellcolor[gray]{0.9}-}  \\\bottomrule\end{tabular}%
		}
	}
\end{table*}
\vspace{-1em}
\subsection{Comparison with the State of the Art}

We evaluate the performance of state-of-the-art deepfake detectors on diffusion-generated images using GenFace. We select various detectors such as CNN-based (i.e., Xception \cite{Xception} and FreqNet \cite{Freq}) , transformer-based (i.e., ViT \cite{ViT}, CViT \cite{CViT}, and FatFormer \cite{FatFormer}) , vision-language-based models (i.e., CLIP \cite{clip}, VLFFD \cite{VLFFD}, and DD-VQA \cite{DD-VQA}), and methods dedicated to detecting diffusion forgery images. (e.g., DIRE \cite{DIRE} and CAEL \cite{genface}). To ensure a fair comparison, we implement the SOTA methods using publicly available codebases and follow the original papers' configurations. All models are trained and evaluated using the same dataset settings. The specific implementation details are as follows:

{\bfseries\setlength\parindent{0em}  Xception and FreqNet.} Since the Xception and FreqNet model codes have been open-sourced, to ensure a fair comparison, we train and test the open-source models using hyperparameter settings in the original paper, and the same dataset protocol as MFCLIP, until convergence, respectively.

{\bfseries\setlength\parindent{0em}  CViT and FatFormer.} We train and test the open-source CViT and FatFormer model using hyperparameter settings in the original paper, and the same dataset protocol as MFCLIP, until convergence, respectively.

{\bfseries\setlength\parindent{0em}  ViT and CLIP.} CLIP tends to be regarded as a foundational vision-language model that serves as a baseline. We use the official OpenAI implementation with the ViT-B/32 backbone. Pre-trained weights are downloaded from the official repository. We train CLIP on our datasets using the Adam optimizer with a learning rate of 1e-5 and weight decay of 1e-4, until convergence. The open-source ViT-B model is trained from scratch on our datasets via the Adam optimizer with a learning rate of 1e-4 and weight decay of 1e-3, until convergence.

{\bfseries\setlength\parindent{0em}  VLFFD and DD-VQA.}  Since VLFFD and DD-VQA do not provide open-source implementations, we follow the descriptions in the original papers, to replicate the models using PyTorch frameworks. To ensure the correctness of our implementation, we compare the performance of replicated models with the results reported in the original papers on standard benchmarks. To ensure a fair comparison, we train and test the replicated model using hyperparameter settings in the original paper, and the same dataset protocol as MFCLIP, until convergence. 

{\bfseries\setlength\parindent{0em}  DIRE and CAEL.} DIRE leverages the discrepancy between an input image and the corresponding reconstruction for diffusion image detection. CAEL integrates the discriminative edges global features with appearance global manipulated representations, to detect diffusion face forgery images. We train and test the CAEL and open-source DIRE model using hyperparameter settings in the original paper, and the same dataset protocol as MFCLIP, until convergence, respectively.

	\begin{figure*}[t!]
	\centering
	\includegraphics[width=\linewidth]{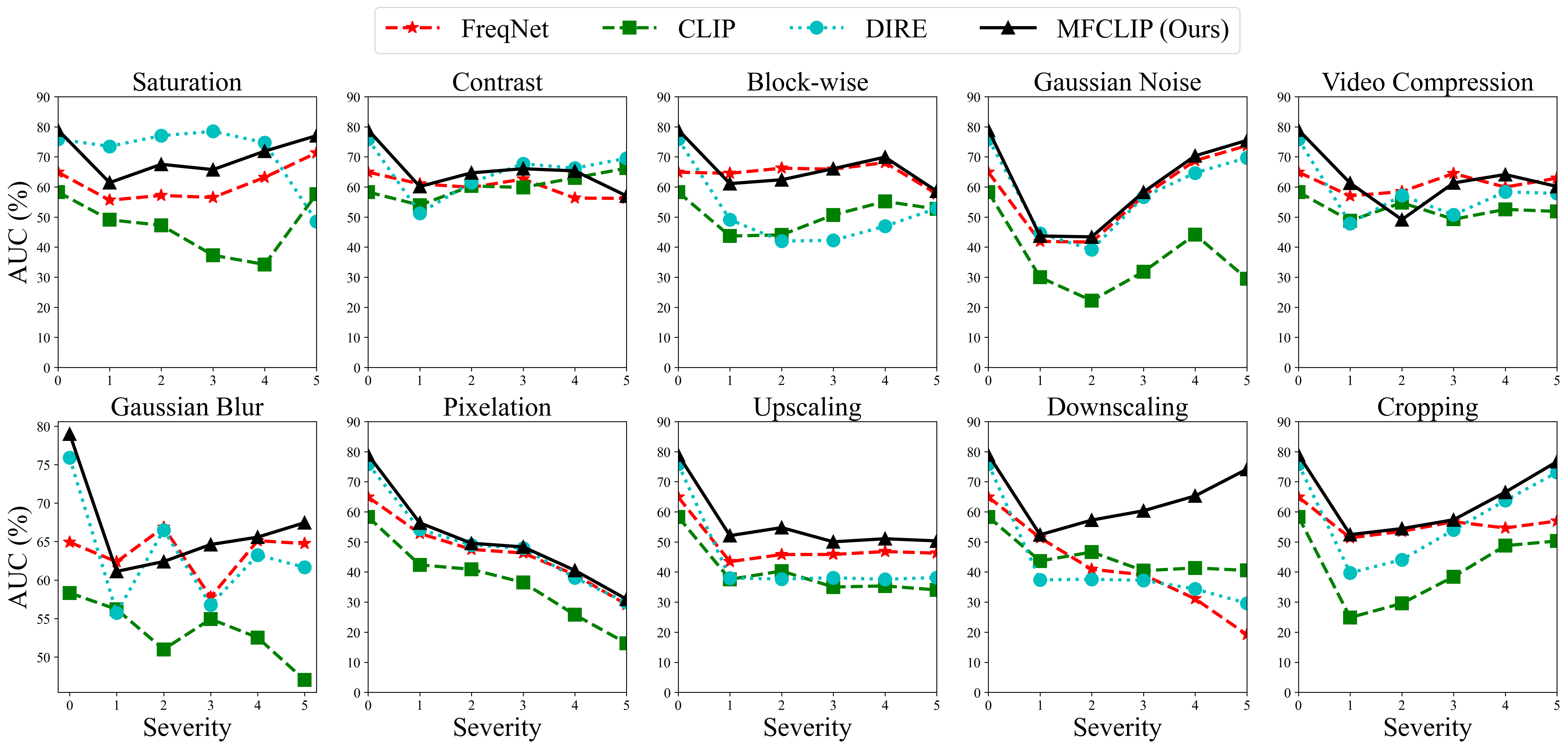} 
	\caption{ The robustness of models to unseen various image perturbations.}\label{figrob}
\end{figure*}

\begin{table}[t!]
	\caption{Cross-generator evaluation on AM. ACC and AUC scores (\%) on remaining generators, after training using one generator in GenFace. 
		\label{tabam}}
	\setlength{\tabcolsep}{1.0mm}{
		\begin{tabular}{rrrrrrrr}\toprule\multicolumn{1}{c}{\multirow{3}[6]{*}{Training Set}} & \multicolumn{1}{c}{\multirow{3}[6]{*}{Model}} & \multicolumn{6}{c}{Testing Set } \\\cmidrule{3-8}      &       & \multicolumn{2}{c}{Diffae} & \multicolumn{2}{c}{LatTrans} & \multicolumn{2}{c}{IAFaces} \\\cmidrule{3-8}      &       & \multicolumn{1}{c}{ACC} & \multicolumn{1}{c}{\cellcolor[gray]{0.9}AUC} & \multicolumn{1}{c}{ACC} & \multicolumn{1}{c}{\cellcolor[gray]{0.9}AUC} & \multicolumn{1}{c}{ACC} & \multicolumn{1}{c}{\cellcolor[gray]{0.9}AUC} \\\midrule\multicolumn{1}{c}{\multirow{7}[2]{*}{Diffae}} & \multicolumn{1}{c}{Xception \cite{Xception}} &    \multicolumn{1}{c}{-}   &    \multicolumn{1}{c}{\cellcolor[gray]{0.9}-}   & \multicolumn{1}{c}{50.00 } & \multicolumn{1}{c}{\cellcolor[gray]{0.9}68.76} & \multicolumn{1}{c}{52.05 } & \multicolumn{1}{c}{\cellcolor[gray]{0.9}55.51} \\      & \multicolumn{1}{c}{ViT \cite{ViT}} &     \multicolumn{1}{c}{-}  &     \multicolumn{1}{c}{\cellcolor[gray]{0.9}-}  & \multicolumn{1}{c}{55.35} & \multicolumn{1}{c}{\cellcolor[gray]{0.9}76.10} & \multicolumn{1}{c}{50.45 } & \multicolumn{1}{c}{\cellcolor[gray]{0.9}54.20 } \\      & \multicolumn{1}{c}{CViT \cite{CViT}} &   \multicolumn{1}{c}{-}    &    \multicolumn{1}{c}{\cellcolor[gray]{0.9}-}  & \multicolumn{1}{c}{50.00 } & \multicolumn{1}{c}{\cellcolor[gray]{0.9}52.03} & \multicolumn{1}{c}{49.95 } & \multicolumn{1}{c}{\cellcolor[gray]{0.9}66.64 } \\      & \multicolumn{1}{c}{DIRE \cite{DIRE}} &    \multicolumn{1}{c}{-}   &    \multicolumn{1}{c}{\cellcolor[gray]{0.9}-}   & \multicolumn{1}{c}{50.00 } & \multicolumn{1}{c}{\cellcolor[gray]{0.9}41.69} & \multicolumn{1}{c}{50.30 } & \multicolumn{1}{c}{\cellcolor[gray]{0.9}63.01 } \\      & \multicolumn{1}{c}{FreqNet \cite{Freq}} &   \multicolumn{1}{c}{-}    &  \multicolumn{1}{c}{\cellcolor[gray]{0.9}-}     & \multicolumn{1}{c}{50.21 } & \multicolumn{1}{c}{\cellcolor[gray]{0.9}51.06} & \multicolumn{1}{c}{50.00 } & \multicolumn{1}{c}{\cellcolor[gray]{0.9}63.91 } \\      & \multicolumn{1}{c}{CLIP \cite{clip}} &    \multicolumn{1}{c}{-}   &  \multicolumn{1}{c}{\cellcolor[gray]{0.9}-}     & \multicolumn{1}{c}{50.00 } & \multicolumn{1}{c}{\cellcolor[gray]{0.9}58.83} & \multicolumn{1}{c}{49.80 } & \multicolumn{1}{c}{\cellcolor[gray]{0.9}49.65 } \\  
			& \multicolumn{1}{c}{FatFormer \cite{FatFormer}} &    \multicolumn{1}{c}{-}   &  \multicolumn{1}{c}{\cellcolor[gray]{0.9}-}     & \multicolumn{1}{c}{50.34 } & \multicolumn{1}{c}{\cellcolor[gray]{0.9}59.67} & \multicolumn{1}{c}{50.93 } & \multicolumn{1}{c}{\cellcolor[gray]{0.9}52.45 } \\  
			& \multicolumn{1}{c}{VLFFD \cite{VLFFD}} &    \multicolumn{1}{c}{-}   &  \multicolumn{1}{c}{\cellcolor[gray]{0.9}-}     & \multicolumn{1}{c}{51.02 } & \multicolumn{1}{c}{\cellcolor[gray]{0.9}60.74} & \multicolumn{1}{c}{51.02 } & \multicolumn{1}{c}{\cellcolor[gray]{0.9}72.81 } \\    
			& \multicolumn{1}{c}{DD-VQA \cite{DD-VQA}} &    \multicolumn{1}{c}{-}   &  \multicolumn{1}{c}{\cellcolor[gray]{0.9}-}     & \multicolumn{1}{c}{51.36 } & \multicolumn{1}{c}{\cellcolor[gray]{0.9}69.04} & \multicolumn{1}{c}{53.84 } & \multicolumn{1}{c}{\cellcolor[gray]{0.9}74.90 } \\    
			& \multicolumn{1}{c}{\textbf{MFCLIP (Ours)}} &     \multicolumn{1}{c}{-}  &   \multicolumn{1}{c}{\cellcolor[gray]{0.9}-}    & \multicolumn{1}{c}{\textbf{59.96} } & \multicolumn{1}{c}{\textbf{\cellcolor[gray]{0.9}79.76}} & \multicolumn{1}{c}{\textbf{54.55 }} & \multicolumn{1}{c}{\textbf{\cellcolor[gray]{0.9}90.27}} \\\midrule\multicolumn{1}{c}{\multirow{7}[2]{*}{LatTrans}} & \multicolumn{1}{c}{Xception \cite{Xception}} & \multicolumn{1}{c}{51.31 } & \multicolumn{1}{c}{\cellcolor[gray]{0.9}63.14 } &  \multicolumn{1}{c}{-}     &  \multicolumn{1}{c}{\cellcolor[gray]{0.9}-}     & \multicolumn{1}{c}{50.05 } & \multicolumn{1}{c}{\cellcolor[gray]{0.9}50.07 } \\      & \multicolumn{1}{c}{ViT \cite{ViT}} & \multicolumn{1}{c}{49.73 } & \multicolumn{1}{c}{\cellcolor[gray]{0.9}49.03 } &     \multicolumn{1}{c}{-}  &      \multicolumn{1}{c}{\cellcolor[gray]{0.9}-} & \multicolumn{1}{c}{50.05 } & \multicolumn{1}{c}{\cellcolor[gray]{0.9}50.97 } \\      & \multicolumn{1}{c}{CViT \cite{CViT}} & \multicolumn{1}{c}{50.76 } & \multicolumn{1}{c}{\cellcolor[gray]{0.9}62.08 } &    \multicolumn{1}{c}{-}   &    \multicolumn{1}{c}{\cellcolor[gray]{0.9}-}   & \multicolumn{1}{c}{50.25 } & \multicolumn{1}{c}{\cellcolor[gray]{0.9}51.71 } \\      & \multicolumn{1}{c}{DIRE \cite{DIRE}} & \multicolumn{1}{c}{50.02 } & \multicolumn{1}{c}{\cellcolor[gray]{0.9}52.35 } &   \multicolumn{1}{c}{-}    & \multicolumn{1}{c}{\cellcolor[gray]{0.9}-}      & \multicolumn{1}{c}{50.00 } & \multicolumn{1}{c}{\cellcolor[gray]{0.9}56.72 } \\      & \multicolumn{1}{c}{FreqNet \cite{Freq}} & \multicolumn{1}{c}{49.88 } & \multicolumn{1}{c}{\cellcolor[gray]{0.9}49.16 } &    \multicolumn{1}{c}{-}   &   \multicolumn{1}{c}{\cellcolor[gray]{0.9}-}    & \multicolumn{1}{c}{50.00 } & \multicolumn{1}{c}{\cellcolor[gray]{0.9}60.38 } \\      & \multicolumn{1}{c}{CLIP \cite{clip}} & \multicolumn{1}{c}{50.02 } & \multicolumn{1}{c}{\cellcolor[gray]{0.9}47.69 } & \multicolumn{1}{c}{-}      &    \multicolumn{1}{c}{\cellcolor[gray]{0.9}-}   & \multicolumn{1}{c}{50.00 } & \multicolumn{1}{c}{\cellcolor[gray]{0.9}53.05 } \\  
			& \multicolumn{1}{c}{FatFormer \cite{FatFormer}} &    \multicolumn{1}{c}{50.05 }   &  \multicolumn{1}{c}{\cellcolor[gray]{0.9}50.21}     & \multicolumn{1}{c}{-} & \multicolumn{1}{c}{\cellcolor[gray]{0.9}-} & \multicolumn{1}{c}{50.09 } & \multicolumn{1}{c}{\cellcolor[gray]{0.9}54.87 } \\    
			& \multicolumn{1}{c}{VLFFD \cite{VLFFD}} &    \multicolumn{1}{c}{50.07}   &  \multicolumn{1}{c}{\cellcolor[gray]{0.9}63.76}     & \multicolumn{1}{c}{- } & \multicolumn{1}{c}{\cellcolor[gray]{0.9}-} & \multicolumn{1}{c}{50.23 } & \multicolumn{1}{c}{\cellcolor[gray]{0.9}65.96 } \\    
			& \multicolumn{1}{c}{DD-VQA \cite{DD-VQA}} &    \multicolumn{1}{c}{52.65}   &  \multicolumn{1}{c}{\cellcolor[gray]{0.9}65.00}     & \multicolumn{1}{c}{- } & \multicolumn{1}{c}{\cellcolor[gray]{0.9}-} & \multicolumn{1}{c}{50.12 } & \multicolumn{1}{c}{\cellcolor[gray]{0.9}66.45 } \\    
			& \multicolumn{1}{c}{\textbf{MFCLIP (Ours)}} & \multicolumn{1}{c}{\textbf{53.00} } & \multicolumn{1}{c}{\cellcolor[gray]{0.9}\textbf{75.00 }} &     \multicolumn{1}{c}{-}  &  \multicolumn{1}{c}{\cellcolor[gray]{0.9}-}     & \multicolumn{1}{c}{\textbf{55.26 }} & \multicolumn{1}{c}{\cellcolor[gray]{0.9}\textbf{75.41} } \\\midrule    \\\end{tabular}%
	}
	\vspace{-3em}
\end{table}

\begin{table}[t]
	\caption{Cross-dataset generalization. * denotes the results of the models are reproduced by ourselves. For the remaining models, we use the results reported in the paper. ACC and AUC scores on FF++, Celeb-DF, DFDC, and DF-1.0 after training using FF++.
		\label{tabcd}}
	\setlength{\tabcolsep}{1.0mm}{ 
		\begin{tabular}{ccccccccc}\toprule
			\multirow{2}[3]{*}{Method} & \multicolumn{2}{c}{FF++} & \multicolumn{2}{c}{Celeb-DF} & \multicolumn{2}{c}{DFDC} & \multicolumn{2}{c}{DF-1.0} \\
			\cmidrule{2-9}      & ACC   & \cellcolor[gray]{0.9}AUC   & ACC   & \cellcolor[gray]{0.9}AUC   & ACC   & \cellcolor[gray]{0.9}AUC   & ACC   & \cellcolor[gray]{0.9}AUC \\\midrule ViT* \cite{ViT}   & 62.44 & \cellcolor[gray]{0.9}67.07 & 62.28 & \cellcolor[gray]{0.9}59.75 & 56.18 & \cellcolor[gray]{0.9}58.31 & 58.05 & \cellcolor[gray]{0.9}61.27 \\CViT* \cite{CViT}  & 90.47 & \cellcolor[gray]{0.9}96.69 & 50.75  & \cellcolor[gray]{0.9}64.70  & 60.95 & \cellcolor[gray]{0.9}65.96 & 56.15 & \cellcolor[gray]{0.9}62.42 \\RECCE \cite{cao2022end} & 97.06 & \cellcolor[gray]{0.9}99.32 &  \multicolumn{1}{c}{-}     & \cellcolor[gray]{0.9}68.71 &   \multicolumn{1}{c}{-}    & \cellcolor[gray]{0.9}69.06 &   \multicolumn{1}{c}{-}    & \multicolumn{1}{c}{\cellcolor[gray]{0.9}-} \\CEViT* \cite{coccomini2022combining} & 93.67 & \cellcolor[gray]{0.9}98.36 & 44.24  & \cellcolor[gray]{0.9}65.29 & 66.14 & \cellcolor[gray]{0.9}75.55 & 62.16 & \cellcolor[gray]{0.9}67.51 \\FoCus \cite{FoCus} & 96.43 & \cellcolor[gray]{0.9}99.15 &    \multicolumn{1}{c}{-}   & \cellcolor[gray]{0.9}76.13 &   \multicolumn{1}{c}{-}    & \cellcolor[gray]{0.9}68.42 &    \multicolumn{1}{c}{-}  &\multicolumn{1}{c}{\cellcolor[gray]{0.9}-}  \\UIA-ViT \cite{UIA-ViT}  &    \multicolumn{1}{c}{-}   & \cellcolor[gray]{0.9}99.33 &   \multicolumn{1}{c}{-}   & \cellcolor[gray]{0.9}82.41 &  \multicolumn{1}{c}{-}     & \cellcolor[gray]{0.9}75.80  &   \multicolumn{1}{c}{-}    & \multicolumn{1}{c}{\cellcolor[gray]{0.9}-} \\Yu et al. \cite{yu}&   \multicolumn{1}{c}{-}    & \cellcolor[gray]{0.9}99.55  &  \multicolumn{1}{c}{-}    & \cellcolor[gray]{0.9}72.86 &   \multicolumn{1}{c}{-}    & \cellcolor[gray]{0.9}69.23 &     \multicolumn{1}{c}{-}  & \multicolumn{1}{c}{\cellcolor[gray]{0.9}-}
			\\Guan et al. \cite{Guan}& - &\cellcolor[gray]{0.9}99.17&  -&\textbf{\cellcolor[gray]{0.9}95.14}& - &\cellcolor[gray]{0.9}74.65& - & \cellcolor[gray]{0.9}- 
			\\CLIP* \cite{clip}  &    67.79   &     \cellcolor[gray]{0.9}69.57  &     64.18  &    \cellcolor[gray]{0.9}65.42   &  58.42     &    \cellcolor[gray]{0.9}57.65   &  57.63     &\cellcolor[gray]{0.9}56.01  \\VLFFD \cite{VLFFD} &   \multicolumn{1}{c}{-}    & \cellcolor[gray]{0.9}99.23 &      \multicolumn{1}{c}{-} & \cellcolor[gray]{0.9}84.80  &    \multicolumn{1}{c}{-}   & \cellcolor[gray]{0.9}84.74  &   \multicolumn{1}{c}{-}    &\multicolumn{1}{c}{\cellcolor[gray]{0.9}-}  \\MFCLIP  &  \textbf{98.15}     &    \textbf{\cellcolor[gray]{0.9}99.63}   &   \textbf{74.02 }   &  \cellcolor[gray]{0.9}83.46     &   \textbf{79.36}    &    \textbf{\cellcolor[gray]{0.9}86.08}   &   \textbf{70.47}    & \textbf{\cellcolor[gray]{0.9}78.99} \\\bottomrule
		\end{tabular}%
	}
\end{table}

\subsubsection{Cross-generator evaluation}

To conduct an in-depth study of the proposed network for generalizable DFFD, we perform the cross-diffusion evaluation on GenFace. Since DFFD is a new and challenging task, to the best of our knowledge, there have been no comprehensive experiments to evaluate the performance of FFD models on cross-diffusion generators. Therefore, our work is the first to assess the generalization of models to unseen diffusion-generated facial images, systematically and comprehensively. Specifically, we train models using the images generated by one diffusion-based generator and test them on different ones. As Table~\ref{tabdiff} shows, detectors tend to achieve better performance on images produced by the generator with high similarity to the generator used for training. Specifically, since DDPM and DiffFace resemble each other, the detector trained using DDPM shows excellent results on images synthesized by DiffFace, and vice versa. For instance, the AUC of Xception, CViT and DIRE is 96.80\%, 96.21\%, and 94.33\% on DiffFace after training using DDPM, respectively. By contrast, they only acquire 63.57\% AUC, 74.51\% AUC, and 62.50\% AUC on LatDiff, individually. 

In addition, as Fig.~\ref{figab} (c) displays, networks generally perform worse on the images generated by LatDiff than on that synthesized by other generators such as DDPM, CollDiff, DiffFace, and Diffae. We believe that facial images produced by LatDiff are realistic, such that the detector struggles to distinguish their authenticity. Therefore, those generated by LatDiff pose a huge threat to the detector. We further perform the cross-generator protocol on AM and FS, respectively. As Table~\ref{tabam} and Table~\ref{tabfs} display, most models acquire poor performance (about 60\% AUC) on the cross-generator evaluation. The AUC of our network is around 40.60\%, 27.26\%, and 26.36\% higher than that of CLIP, DIRE, and FreqNet, respectively, on IAFaces after training using Diffae. Besides, the AUC of our MFCLIP model is about 24\% higher than that of CAEL on DDPM after training using CollDiff. We believe that our MFCLIP model can extract more fine-grained language representations and conduct vision-language contrastive learning, adaptively, thus obtaining excellent performance.

\subsubsection{ Cross-forgery evaluation} 
To investigate the generalization of various detectors, we perform cross-forgery tests. We train models using images of one manipulation, and test them on those of remaining manipulations. As Table~\ref{tab1} shows, the performance of our methods outperforms most detectors, demonstrating the superior generalization of MFCLIP. Specifically, for the vision-language-based models, the AUC of our model is nearly 26.80\%, 18.78\%, and 14.66\% higher than that of CLIP, VLFFD, and DD-VQA, respectively, on EFS after training using AM, which is attributed that our MFCLIP model not only captures global visual forgery patterns across image noise modalities but also achieves better vision-language alignment.

\subsubsection{ Cross-dataset evaluation} 
To investigate the generalization of MFCLIP, we conduct cross-dataset evaluations. We train detectors using FF++ and test them on FF++, CelebDF, DFDC, and DF-1.0. The first two level text prompts in FF++ are only introduced to train MFCLIP, due to the limitation of labels offered by FF++. As Table~\ref{tabcd} shows, the AUC of our MFCLIP method is about 1.34\% higher than that of VLFFD. We argue that our MFCLIP model can capture discriminative global noise features, and achieve more flexible vision-language alignment, compared to the VLFFD model.

\subsubsection{Generalization to unseen generative models} To further study the generalization of different detectors to unseen generative models, we train networks using images generated by various generators from GenFace, and test them on those synthesized by autoregressive, VAE-based, and flow-based approaches. In particular, we leverage autoregressive models (e.g. HART \cite{hart} and VAR \cite{var}), VAE-based generators (e.g. vanilla VAE \cite{vae}), and flow-based generative networks (e.g. FLUX \cite{flux} and FlowEdit \cite{flowedit}), to generate 2k face manipulated images, respectively. We also choose 2k fake face images created by the VAE-based DF-VAE \cite{DF1.0} model from the DF-1.0 dataset. We evaluate the performance of various detectors on these face forgery images, accordingly. In Table~\ref{flow}, methods tend to achieve lower ACC scores on face images created by VAR than on those generated by HART. This indicates that VAR which shows SOTA performance creates high-quality face images that are more complex and challenging. To further improve generalization to different forgery types, we introduce various generator distributions under level 4 in GenFace to train the detector, to capture more general forgery features. Specifically, we arbitrarily adopt 45k real face images and 45k fake ones from GenFace, with the fake images evenly distributed across ten different GAN or diffusion generators under level 4 (4.5k images per generator). To ensure a fair comparison, all detectors are trained using the same generator distributions from GenFace. In Table~\ref{flow}, the AUC of MFCLIP is around 88\% on flow-based models. Compared to the state-of-the-art DIRE, MFCLIP achieves superior performance on images created by flow-based generators. 
	
		Despite the better generalization to different forgery types than other detectors, we need to improve transferability across diverse generators such as autoregressive, VAE-based, and flow-based methods. In the future, we aim to explore discriminative prior characteristics of various forgery types, such as noise patterns in flow-based methods or reconstruction errors in autoregressive models, and incorporate them into the detection model. Besides, multi-modal large language models could be integrated with our detector to capture more complex, nuanced, and general patterns in forgery face data.

\subsubsection{ Robustness to common image corruptions} 
We assess the robustness of detectors against different unseen image distortions. We train models on GenFace and test their performance on distorted images from \cite{DF1.0}. Ten types of perturbations are involved, each with five intensity levels. As shown in Fig.~\ref{figrob}, we test the models on various image distortions, such as saturation changes, contrast adjustments, block distortions, white Gaussian noise, blurring, pixelation, video compression, upscaling, downscaling, and cropping.  An intensity of 0 indicates no degradation.
	
Specifically, the specific settings for the severity of the first seven deformations are discussed in detail in \cite{genface}. Upscaling aims to enlarge the image pixel dimensions using scaling factors [1.1, 1.3, 1.5, 2.0, 3.0] with five intensities (the larger the factor value, the greater the upscaling rate). Downscaling intends to reduce the image pixel dimensions by applying scaling parameters such as [0.91, 0.77, 0.67, 0.50, 0.33], which correspond to five intensities of downscaling (the smaller the parameter value, the greater the downscaling degree). Cropping removes a portion of the image based on crop percentages [0.8, 0.7, 0.6, 0.5, 0.4], with each percentage representing a different level of image retention (the smaller the percentage, the greater the cropping severity). When adopting perturbations of different severities, the changes in AUC for all detectors are presented in Fig.~\ref{figrob}. We observe that the AUC of models generally decreases with the increase of downscaling intensities. By contrast, the AUC of our MFCLIP method shows an increased trend, which demonstrates the strong robustness of MFCLIP. Besides, the performance of detectors reaches the minimum when the severity of cropping equals one, and begins to rise thereafter. This trend indicates that detectors could generalize well to smaller facial areas when trained using whole face images. The results show that our model surpasses most methods across various types of image degradation.
\begin{figure*}[t!]
	\centering 
	\includegraphics[width=\linewidth]{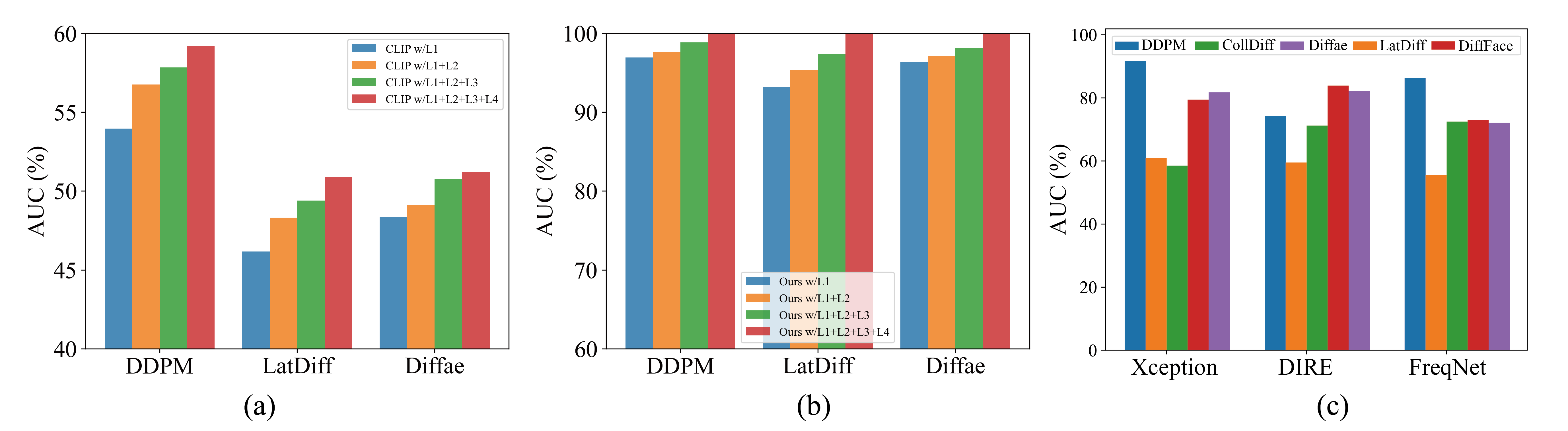} 
	\caption{(a) and (b) display the ablation results of CLIP and MFCLIP with hierarchical fine-grained text prompts, respectively. (c) shows the comparison of the performance of detectors on various generators. We average the AUC scores of the model listed in Table~\ref{tabdiff} on the images from the same generator. }\label{figab}
\end{figure*}

\begin{table}[t!]
	\caption{MFCLIP ablation. We test models on DDPM, LatDiff, DiffFace, and Diffae, after training using CollDiff in GenFace. \label{tababla}}
	
	\setlength{\tabcolsep}{0.45mm}{
		\begin{tabular}{ccccccccccccc}
			\toprule
			\multicolumn{5}{c}{\multirow{2}[4]{*}{Model}} &
			\multicolumn{8}{c}{Testing Set} \\
			\cmidrule(lr){6-13}
			& \multicolumn{4}{c}{} &
			\multicolumn{2}{c}{DDPM} &
			\multicolumn{2}{c}{LatDiff} &
			\multicolumn{2}{c}{DiffFace} &
			\multicolumn{2}{c}{Diffae} \\
			\midrule
			NE & IE & FLE & Predictor & SPA &
			ACC & \cellcolor[gray]{0.9}AUC &
			ACC & \cellcolor[gray]{0.9}AUC &
			ACC & \cellcolor[gray]{0.9}AUC &
			ACC & \cellcolor[gray]{0.9}AUC \\
			\midrule
			\checkmark & \multicolumn{1}{c}{-} & \multicolumn{1}{c}{-} & \multicolumn{1}{c}{-} & \multicolumn{1}{c}{-} &
			99.62 & \cellcolor[gray]{0.9}98.83 &
			88.78 & \cellcolor[gray]{0.9}88.60 &
			91.06 & \cellcolor[gray]{0.9}98.93 &
			89.79 & \cellcolor[gray]{0.9}99.86 \\
			\multicolumn{1}{c}{-} & \checkmark & \multicolumn{1}{c}{-} & \multicolumn{1}{c}{-} & \multicolumn{1}{c}{-} &
			99.74& \cellcolor[gray]{0.9}99.97 &
			49.98 & \cellcolor[gray]{0.9}47.59 &
			81.97& \cellcolor[gray]{0.9}98.80 &
			90.56 & \cellcolor[gray]{0.9}99.83 \\
			\checkmark & \checkmark & \multicolumn{1}{c}{-} & \multicolumn{1}{c}{-} & \multicolumn{1}{c}{-} &
			99.77 & \cellcolor[gray]{0.9}99.99 &
			92.77 & \cellcolor[gray]{0.9}95.48 &
			92.56 & \cellcolor[gray]{0.9}99.77 &
			92.35 & \cellcolor[gray]{0.9}99.86\\
			\multicolumn{1}{c}{-} & \checkmark & \checkmark & \multicolumn{1}{c}{-} & \multicolumn{1}{c}{-} &
			99.76 & \cellcolor[gray]{0.9}99.99 &
			92.00 & \cellcolor[gray]{0.9}93.14 &
			92.45 & \cellcolor[gray]{0.9}99.86 &
			92.11 & \cellcolor[gray]{0.9}99.96 \\
			\multicolumn{1}{c}{-} & \checkmark & \checkmark & \checkmark & \multicolumn{1}{c}{-} &
			99.98 & \cellcolor[gray]{0.9}99.99 &
			92.07 & \cellcolor[gray]{0.9}95.04 &
			95.48 & \cellcolor[gray]{0.9}99.93 &
			92.17 & \cellcolor[gray]{0.9}99.97 \\
			\checkmark & \checkmark & \checkmark & \checkmark & \multicolumn{1}{c}{-} &
			99.99 & \cellcolor[gray]{0.9}99.99 &
			93.42 & \cellcolor[gray]{0.9}96.18 &
			98.08& \cellcolor[gray]{0.9}99.87 &
			92.18 & \cellcolor[gray]{0.9}99.98 \\
			\checkmark & \checkmark & \checkmark & \checkmark & \checkmark &
			100 & \cellcolor[gray]{0.9}100 &
			99.19 & \cellcolor[gray]{0.9}99.97 &
			99.63 & \cellcolor[gray]{0.9}99.96 &
			93.94 & \cellcolor[gray]{0.9}99.99 \\
			\bottomrule
		\end{tabular}
	}
\end{table}
\vspace{-1em}
\section{Ablation Study}\label{sec5}
We conduct an ablation study to evaluate the contribution of each component in MFCLIP. In this study, we use images generated by five diffusion-based models including CollDiff, DDPM, LatDiff, DiffFace and Diffae, to investigate the effectiveness of MFCLIP. We consider six schemes: 1) the impacts of components, 2) the influence of sample pair attention, 3) the effects of fine-grained text prompts, 4) the influences of loss functions, 5) the effect of the patch size in PS, 6) Impacts of pre-trained weights of CLIP, and 7) Impacts of the FID distribution in FTG. In the subsequent subsection, we discuss the seven aspects, respectively.

\begin{table*}[t!]
	\caption{ The FID scores of various forgery categories under different levels in FTG. We compute the FID scores on the GenFace dataset. The lower FID scores indicate better image quality.  \label{fid}}

\setlength{\tabcolsep}{1.22mm}{
\begin{tabular}{c|cccccccccccc}
	\toprule
	\multirow{2}[4]{*}{} & \multicolumn{1}{c|}{L1} & \multicolumn{4}{c|}{L2}       & \multicolumn{7}{c}{L3} \\
	\cmidrule{2-13}      & \multicolumn{1}{c|}{Fake} & EFS   & FS    & AM    & \multicolumn{1}{c|}{Average} & EFS GAN & EFS Diffusion & FS GAN & FS Diffusion & AM GAN & AM Diffusion & Average \\
	\cmidrule{1-13}FID   & \multicolumn{1}{c|}{13.097} & 13.353 & 17.145 & 31.257 & \multicolumn{1}{c|}{20.585} & 40.041 & 12.368 & 21.556 & 13.908 & 44.349 & 42.253 & 29.079 \\
	\midrule
	\multirow{2}[4]{*}{} & \multicolumn{12}{c}{L4} \\
	\cmidrule{2-13}      & StyleGAN2 & StyleGAN3 & CollDiff & DDPM  & LatDiff & FSLSD & FaceSwapper & DiffFace & Diffae & LatTrans & \multicolumn{2}{c}{Average} \\
	\midrule
	FID   & 39.102 & 42.654 & 20.855 & 35.738 & 11.274 & 6.517 & 50.738 & 13.908 & 42.253 & 44.349 & \multicolumn{2}{c}{30.739} \\
	\bottomrule
\end{tabular}%
}
\end{table*}
\vspace{-1.5em}
\subsection{Impacts of components}

To examine the contribution of each component to learning ability, we observe the performance of models on DDPM, LatDiff, DiffFace, and Diffae after training using CollDiff. Table~\ref{tababla} shows the ablation results of the model. NE improves the performance by 47.89\% AUC on LatDiff, confirming that noises extracted from the richest patch offer valuable information to benefit the DFFD. The gains from introducing the FLE module (+45.55\%) are obvious, demonstrating the significance of fine-grained text embeddings. Predictor increases performance by 3.23\% ACC on DiffFace, showing that DFFD could benefit from aligning visual and linguistic representations in feature space. The introduction of SPA further enhances the performance (+3.79\%), which verifies that cross-modality sample pairs are adaptively emphasized and suppressed, guiding the model to achieve better feature alignment.

\begin{table}[t!]
	\caption{Ablation results of various patch sizes. We test models on DDPM, LatDiff, DiffFace, and Diffae, after training on CollDiff in GenFace. \label{tabps}}
	\small
	\setlength{\tabcolsep}{1.0mm}{
		\begin{tabular}{ccccccccc}
			\toprule
			\multirow{2}[3]{*}{Patch Size} & \multicolumn{2}{c}{DDPM} & \multicolumn{2}{c}{LatDiff} & \multicolumn{2}{c}{DiffFace} & \multicolumn{2}{c}{Diffae} \\\cmidrule{2-9}      & ACC   & \cellcolor[gray]{0.9}AUC   & ACC   & \cellcolor[gray]{0.9}AUC   & ACC   & \cellcolor[gray]{0.9}AUC   & ACC   & \cellcolor[gray]{0.9}AUC \\\midrule16    &     56.21  &    \cellcolor[gray]{0.9}81.69   &  50.00     &    \cellcolor[gray]{0.9}41.57   &  50.42     &   \cellcolor[gray]{0.9}68.81    &    57.31   & \cellcolor[gray]{0.9}87.46 \\28    & 80.89 & \cellcolor[gray]{0.9}99.76 & 50.00  & \cellcolor[gray]{0.9}42.69 & 50.49 & \cellcolor[gray]{0.9}75.34 & 53.95 & \cellcolor[gray]{0.9}87.94 \\32    & 98.42 & \cellcolor[gray]{0.9}99.96 & 50.04 & \cellcolor[gray]{0.9}49.62  & 64.49 & \cellcolor[gray]{0.9}93.48 & 63.27 & \cellcolor[gray]{0.9}89.95 \\56    & 56.30  & \cellcolor[gray]{0.9}97.49 & 50.00    & \cellcolor[gray]{0.9}90.79 & 53.50  & \cellcolor[gray]{0.9}97.49  & 51.59 & \cellcolor[gray]{0.9}96.90 \\112   & \textbf{100}   & \textbf{\cellcolor[gray]{0.9}100}   & \textbf{99.19} & \textbf{\cellcolor[gray]{0.9}99.97} & \textbf{99.63} & \textbf{\cellcolor[gray]{0.9}99.96} & \textbf{93.94}& \textbf{\cellcolor[gray]{0.9}99.99} \\224   & 100   & \cellcolor[gray]{0.9}100   & 95.30 & \cellcolor[gray]{0.9}99.83 & 99.60 & \cellcolor[gray]{0.9}99.95 & 65.50 & \cellcolor[gray]{0.9}92.03 \\
			\bottomrule
		\end{tabular}%
	}
\end{table}

\begin{table}[t!]
	\caption{Generalization to various generative models. We test detectors on autoregressive, VAE-based, and flow-based generative methods after training using GenFace. \label{flow}}
	\small
	\setlength{\tabcolsep}{0.8mm}{
\begin{tabular}{ccccccccc}
	\toprule
	\multirow{3}[6]{*}{Method} & \multicolumn{4}{c}{Autoregressive} & \multicolumn{2}{c}{\multirow{2}[4]{*}{VAE-based}} & \multicolumn{2}{c}{\multirow{2}[4]{*}{Flow-based}} \\
	\cmidrule{2-5}      & \multicolumn{2}{c}{HART} & \multicolumn{2}{c}{VAR} & \multicolumn{2}{c}{} & \multicolumn{2}{c}{} \\
	\cmidrule{2-9}      & ACC   & \cellcolor[gray]{0.9}AUC   & ACC   & \cellcolor[gray]{0.9}AUC   & ACC   & \cellcolor[gray]{0.9}AUC   & ACC   & \cellcolor[gray]{0.9}AUC \\
	\midrule
	CViT \cite{CViT}  & 55.50  & \cellcolor[gray]{0.9}87.87 &54.08 & \multicolumn{1}{c}{\cellcolor[gray]{0.9}83.88} & 50.13 & \multicolumn{1}{c}{\cellcolor[gray]{0.9}65.79} & 56.05 & \multicolumn{1}{c}{\cellcolor[gray]{0.9}84.31} \\
	FreqNet \cite{Freq} & 56.10 & \cellcolor[gray]{0.9}84.52 & 52.75 & \multicolumn{1}{c}{\cellcolor[gray]{0.9}77.41} & 52.63 & \multicolumn{1}{c}{\cellcolor[gray]{0.9}72.82} & 62.72 &\multicolumn{1}{c}{\cellcolor[gray]{0.9}87.75} \\
	DIRE \cite{DIRE}  & 61.87  & \cellcolor[gray]{0.9}87.69 & 58.26 & \multicolumn{1}{c}{\cellcolor[gray]{0.9}85.94} & 57.64 & \multicolumn{1}{c}{\cellcolor[gray]{0.9}76.80} & 62.49  & \multicolumn{1}{c}{\cellcolor[gray]{0.9}85.76} \\
	CAEL \cite{genface}  & 54.55 & \cellcolor[gray]{0.9}83.77  & 56.99 &\multicolumn{1}{c}{\cellcolor[gray]{0.9}81.14} & 53.62 & \multicolumn{1}{c}{\cellcolor[gray]{0.9}76.67} & 61.91 & \multicolumn{1}{c}{\cellcolor[gray]{0.9}86.04} \\
	MFCLIP & \textbf{62.74}  &\textbf{\cellcolor[gray]{0.9}88.24} & \textbf{60.32} & \multicolumn{1}{c}{\textbf{\cellcolor[gray]{0.9}86.35}} & \textbf{60.94} &  \textbf{\cellcolor[gray]{0.9}77.31}     &   \textbf{64.90}    &\multicolumn{1}{c}{\textbf{\cellcolor[gray]{0.9}88.19 }}\\
	\bottomrule
\end{tabular}%

}
\end{table}
\vspace{-1em}
\subsection{Effect of the patch size in PS}
We investigate the impact of different patch sizes in PS. The proposed MFCLIP model is trained using CollDiff and tested on DDPM, LatDiff, DiffFace, and Diffae. We report the performance of MFCLIP from patch size 16 to 224. Table~\ref{tabps} shows that the performance typically increases with the growth of patch size. The AUC reaches the peak as the patch size is 112, and begins to decline afterward. We argue that the model tends to explore more forgery areas, when the larger patch size is leveraged. However, oversized patch scales like 224 may introduce noises, enabling the model to acquire poor generalization performance.
\begin{table}[t!]
	\caption{The effect of the SPA module. We test models on DDPM, LatDiff, and Diffae, after training using CollDiff in GenFace. \label{tablespa}}
	\setlength{\tabcolsep}{0.6mm}{
		\begin{tabular}{lcccccccccc}\toprule\multicolumn{1}{c}{\multirow{2}[4]{*}{Method}} & \multicolumn{2}{c}{DDPM} & \multicolumn{2}{c}{LatDiff} & \multicolumn{2}{c}{Diffae} & \multirow{2}[4]{*}{\shortstack{Params \\ (M)} } & \multirow{2}[4]{*}{\shortstack{FLOPs \\ (G)} } \\\cmidrule{2-7}      & ACC   & \cellcolor[gray]{0.9}AUC   & ACC   & \cellcolor[gray]{0.9}AUC     & ACC   & \cellcolor[gray]{0.9}AUC   &       &  \\\midrule CLIP w/o SPA & 51.58 & \cellcolor[gray]{0.9}53.92 & 50.21  & \cellcolor[gray]{0.9}46.17  & 49.82 & \cellcolor[gray]{0.9}48.36 & 84.225 & 117.23 \\CLIP w/ SPA & 52.68 & \cellcolor[gray]{0.9}59.20  & 50.37 & \cellcolor[gray]{0.9}48.89  & 49.98 & \cellcolor[gray]{0.9}50.21 & 84.225 & 117.23 \\MFCLIP w/o SPA & 99.99 & \cellcolor[gray]{0.9}99.99 & 93.42 & \cellcolor[gray]{0.9}96.18  & 92.18 & \cellcolor[gray]{0.9}99.98 & 93.834 & 358.12 \\MFCLIP w/ SPA & \textbf{100} &\textbf{ \cellcolor[gray]{0.9}100} & \textbf{99.19} & \textbf{\cellcolor[gray]{0.9}99.97}& \textbf{93.94} & \textbf{\cellcolor[gray]{0.9}99.99 }& 93.834 & 358.12 \\\bottomrule\end{tabular}%
	}
\end{table}

\begin{figure*}[t!]
	\centering
	\includegraphics[width=\linewidth]{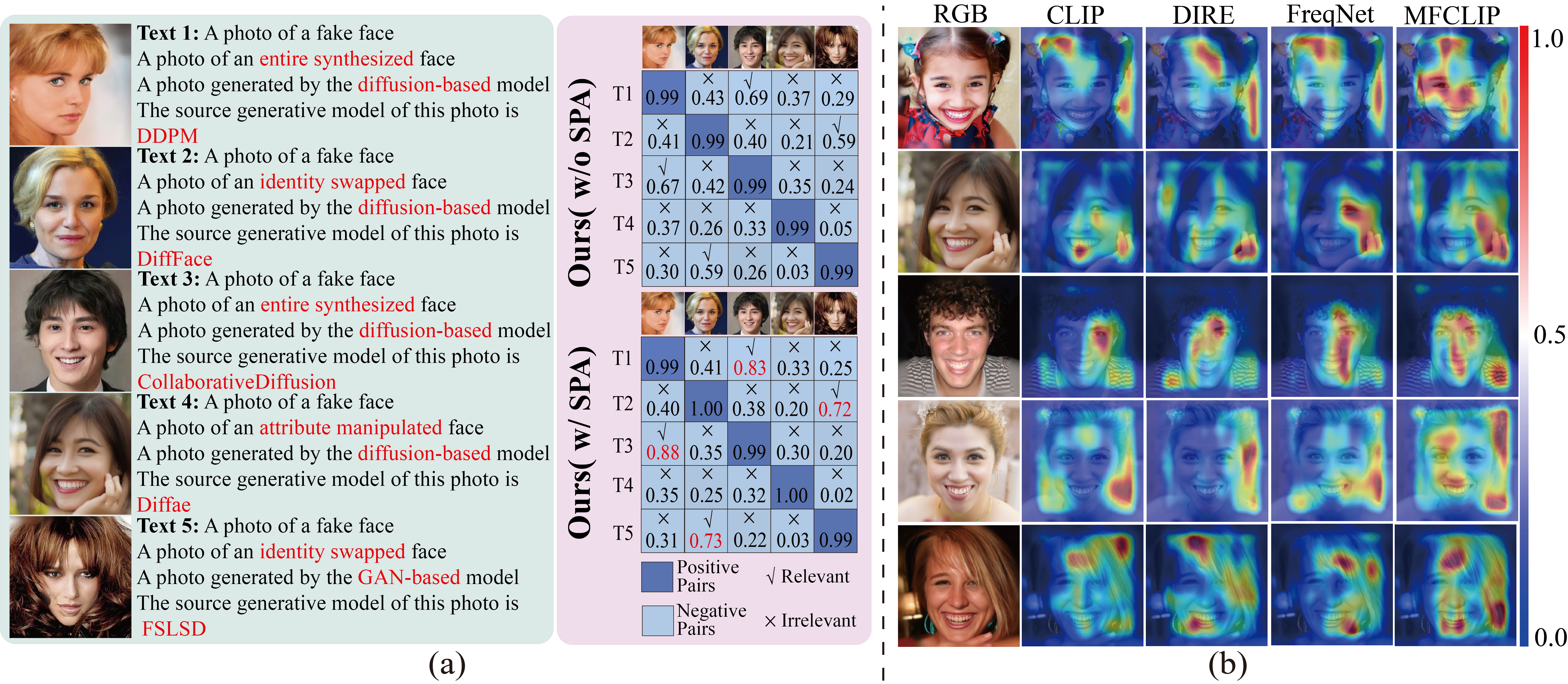} 
	\caption{ (a) The visualization of the cosine similarity matrix in MFCLIP (w/ SPA and w/o SPA). (b) The heatmap visualizations of various detectors on some examples from GenFace.  }\label{figspa}
\end{figure*}
\vspace{-1em}
\subsection{Influence of sample pair attention}
\vspace{-0.2em}
 To evaluate the effectiveness and efficiency of SPA, we conduct the cross-generator evaluation to show that our plug-and-play SPA module can generalize to various vison-language-based models for performance improvement. We train models using CollDiff and test them on DDPM, LatDiff, and Diffae. As Table~\ref{tablespa} shows, due to the addition of SPA, the AUC of both CLIP and MFCLIP is improved by 2.72\% and 3.78\%, respectively, on the challenging LatDiff, with almost no change in the number of parameters and FLOPs. SPA rarely introduces auxiliary weights and computational costs when plugged and played into other models like CLIP. This shows that performance improvements could be achieved with high computational efficiency as the addition of SPA.

\begin{table}[t!]
	\caption{The impact of the fine-grained text prompts. * denotes wider text prompts in the hierarchical structure. L5 means that we introduce the learnable text prompts to build the deeper hierarchical structure. We test models on DDPM, LatDiff, and Diffae, after training using CollDiff in GenFace.\label{tabtext}}
	\setlength{\tabcolsep}{0.25mm}{
\begin{tabular}{rrrrrrrrr}\toprule\multicolumn{1}{c}{\multirow{2}[4]{*}{Method}} & \multicolumn{2}{c}{DDPM} & \multicolumn{2}{c}{LatDiff} & \multicolumn{2}{c}{Diffae} & \multicolumn{1}{c}{\multirow{2}[4]{*}{\hspace{-0.3em}\shortstack{Params \\ (M)} }} & \multicolumn{1}{c}{\multirow{2}[4]{*}{\hspace{-0.35em}\shortstack{ FLOPs\\ (G)} }} \\
	\cmidrule{2-7}      & \multicolumn{1}{c}{\hspace{-0.5em}ACC} & \multicolumn{1}{c}{\cellcolor[gray]{0.9}AUC} & \multicolumn{1}{c}{ACC} & \multicolumn{1}{c}{\cellcolor[gray]{0.9}AUC} & \multicolumn{1}{c}{ACC} & \multicolumn{1}{c}{\cellcolor[gray]{0.9}AUC} &       &  \\
	\midrule\multicolumn{1}{l}{Ours w/L1} & \multicolumn{1}{c}{\hspace{-0.5em}99.01} & \multicolumn{1}{c}{\cellcolor[gray]{0.9}99.04} & \multicolumn{1}{c}{95.82 } & \multicolumn{1}{c}{\cellcolor[gray]{0.9}96.03} & \multicolumn{1}{c}{90.45} & \multicolumn{1}{c}{\cellcolor[gray]{0.9}97.05} & \multicolumn{1}{c}{93.83} & \multicolumn{1}{c}{218.60 } \\\multicolumn{1}{l}{Ours  w/L1+L2} & \multicolumn{1}{c}{\hspace{-0.5em}99.25} & \multicolumn{1}{c}{\cellcolor[gray]{0.9}99.38 } & \multicolumn{1}{c}{96.38} & \multicolumn{1}{c}{\cellcolor[gray]{0.9}96.57} & \multicolumn{1}{c}{90.72} & \multicolumn{1}{c}{\cellcolor[gray]{0.9}97.34} & \multicolumn{1}{c}{93.83} & \multicolumn{1}{c}{265.11} \\\multicolumn{1}{l}{Ours  w/L1+L2+L3} & \multicolumn{1}{c}{\hspace{-0.5em}99.66} & \multicolumn{1}{c}{\cellcolor[gray]{0.9}99.79} & \multicolumn{1}{c}{97.03} & \multicolumn{1}{c}{\cellcolor[gray]{0.9}97.36} & \multicolumn{1}{c}{91.08} & \multicolumn{1}{c}{\cellcolor[gray]{0.9}97.63} & \multicolumn{1}{c}{93.83} & \multicolumn{1}{c}{311.62} \\\multicolumn{1}{l}{Ours  w/L1+L2+L3+L4} & \multicolumn{1}{c}{\hspace{-0.1em}\textbf{100}} & \multicolumn{1}{c}{\cellcolor[gray]{0.9}\textbf{100}} & \multicolumn{1}{c}{\textbf{99.19}} & \multicolumn{1}{c}{\cellcolor[gray]{0.9}\textbf{99.97}} & \multicolumn{1}{c}{93.94} & \multicolumn{1}{c}{\cellcolor[gray]{0.9}\textbf{99.99}} & \multicolumn{1}{c}{93.83} & \multicolumn{1}{c}{358.12} \\\multicolumn{1}{l}{Ours  w/L1*+L2+L3+L4} & \multicolumn{1}{c}{\hspace{-0.5em}99.98} & \multicolumn{1}{c}{\cellcolor[gray]{0.9}100} & \multicolumn{1}{c}{96.48} & \multicolumn{1}{c}{\cellcolor[gray]{0.9}99.79} & \multicolumn{1}{c}{95.82} & \multicolumn{1}{c}{\cellcolor[gray]{0.9}99.57} & \multicolumn{1}{c}{93.83} & \multicolumn{1}{c}{358.12} \\\multicolumn{1}{l}{Ours  w/L1+L2+L3+L4+L5} & \multicolumn{1}{c}{\hspace{-0.1em}100} & \multicolumn{1}{c}{\cellcolor[gray]{0.9}100} & \multicolumn{1}{c}{98.92} & \multicolumn{1}{c}{\cellcolor[gray]{0.9}99.66} & \multicolumn{1}{c}{\textbf{98.44}} & \multicolumn{1}{c}{\cellcolor[gray]{0.9}99.40} & \multicolumn{1}{c}{93.83} & \multicolumn{1}{c}{404.63} \\
		\bottomrule
\end{tabular}%
	}
\end{table}
\vspace{-1em}
\subsection{Effects of fine-grained text prompts}
 To investigate the impact of the hierarchical fine-grained text prompts, we evaluate the performance of CLIP and our MFCLIP method on DDPM, LatDiff, and Diffae after training on CollDiff, by gradually introducing the hierarchical texts. In Fig.~\ref{figab}, we notice that the performance of our model commonly improves with the increase of hierarchical text prompts. Specifically, the AUC of CLIP and our model is improved on DDPM, LatDiff, and Diffae, respectively, as fine-grained texts from levels 1 to 4 are introduced, showing that hierarchical text prompts guide the model to capture more general and discriminative information, to facilitate the advancement of DFFD. 
 
Additionally, Table~\ref{tabtext} shows the performance of MFCLIP with wider or deeper text prompts in the hierarchical structure. In detail, at level 1, we generate detailed text prompts such as ``a photo of a real face with natural skin texture, facial symmetry with natural variation, life-like expressions, and dynamic facial contours" or ``a photo of a fake face with unrealistic skin texture, perfect symmetry without variation, unnatural facial proportions, lack of life-like expressions, inconsistent lighting and shadows, and a flat, two-dimensional appearance", to create a wider hierarchical structure. We also formulate learnable text prompts at level 5, to build a deeper one. We observe that using deeper or wider text prompts in the hierarchical structure struggles to improve the performance of MFCLIP, as they lead to information redundancy, weakening the representation capability. Besides, MFCLIP achieves excellent results while maintaining the trade-off between computational efficiency and performance, when four levels of text prompts are involved. Specifically, an approximately 2\% increase of AUC could be achieved by adding a level of text prompts from level 3 to 4, with no change in the number of parameters but an increase of 47.5G FLOPs. The AUC shows a decreased trend when the level of text prompts equals five.

\begin{table}[t!]
	\caption{Effects of losses. We test models on DDPM, LatDiff, DiffFace, and Diffae, after training on CollDiff in GenFace. \label{tabloss}}
	\setlength{\tabcolsep}{1.1mm}{
		\begin{tabular}{ccccccccc}
			\toprule
			\multicolumn{1}{c}{\multirow{2}[3]{*}{\shortstack{Loss \\ Function}}} & \multicolumn{2}{c}{DDPM} & \multicolumn{2}{c}{LatDiff} & \multicolumn{2}{c}{DiffFace} & \multicolumn{2}{c}{Diffae} \\\cmidrule{2-9}      & ACC   & \cellcolor[gray]{0.9}AUC   & ACC   & \cellcolor[gray]{0.9}AUC   & ACC   & \cellcolor[gray]{0.9}AUC   & ACC   & \cellcolor[gray]{0.9}AUC \\
			\midrule
				\multicolumn{1}{l}{	$\mathcal{L}_\text{ce}$} & 99.77 & \cellcolor[gray]{0.9}99.99 & 92.77  & \cellcolor[gray]{0.9}95.48 & 92.56 & \cellcolor[gray]{0.9}99.77 & 92.35 & \cellcolor[gray]{0.9}99.86 \\\multicolumn{1}{l}{	$\mathcal{L}_\text{ce}$+$\mathcal{L}_\text{kl}$ } &   99.89    &    \cellcolor[gray]{0.9}99.99   &    94.07   &   \cellcolor[gray]{0.9}97.26    &    94.53   &   \cellcolor[gray]{0.9}99.80    &   92.91    &  \cellcolor[gray]{0.9}99.92 \\\multicolumn{1}{l}{$\mathcal{L}_\text{ce}$+$\mathcal{L}_\text{cmc}$} &     99.92  &     \cellcolor[gray]{0.9}99.99  &     95.36  & \cellcolor[gray]{0.9}98.03    &  95.88     & \cellcolor[gray]{0.9}99.85  &     92.98  & \cellcolor[gray]{0.9}99.95\\
			\multicolumn{1}{l}{	$\mathcal{L}_\text{ce}$+$\mathcal{L}_\text{cmc}$+$\mathcal{L}_\text{kl}$}& \textbf{100}   & \textbf{\cellcolor[gray]{0.9}100}   & \textbf{99.19} & \textbf{\cellcolor[gray]{0.9}99.97} & \textbf{99.63} & \textbf{\cellcolor[gray]{0.9}99.96} & \textbf{93.94} & \textbf{\cellcolor[gray]{0.9}99.99} \\
			\bottomrule
		\end{tabular}
	}
\end{table}

\begin{table}[t!]
	\caption{Effects of various generator distributions. We test models on DDPM, DiffFace, and Diffae, after training on LatDiff or both LatDiff and StyleGAN3. SG3 is StyleGAN3. \label{data}}
	\setlength{\tabcolsep}{0.23mm}{
\begin{tabular}{lcrrrrrrrr}
	\toprule
	\multicolumn{1}{c}{\multirow{3}[6]{*}{Training Set}} & \multirow{3}[6]{*}{Model} & \multicolumn{8}{c}{Testing Set } \\
	\cmidrule{3-10}      &       & \multicolumn{2}{c}{DDPM} & \multicolumn{2}{c}{CollDiff} & \multicolumn{2}{c}{DiffFace} & \multicolumn{2}{c}{Diffae} \\
	\cmidrule{3-10}      &       & \multicolumn{1}{c}{ACC} & \multicolumn{1}{c}{\cellcolor[gray]{0.9}AUC} & \multicolumn{1}{c}{ACC} & \multicolumn{1}{c}{\cellcolor[gray]{0.9}AUC} & \multicolumn{1}{c}{ACC} & \multicolumn{1}{c}{\cellcolor[gray]{0.9}AUC} & \multicolumn{1}{c}{ACC} & \multicolumn{1}{c}{\cellcolor[gray]{0.9}AUC} \\
	\midrule

	\hspace{1.1em}LatDiff & Xception \cite{Xception} &   50.13    &    \cellcolor[gray]{0.9}76.45   &   50.00    &    \cellcolor[gray]{0.9}37.05   &   51.51    &    \cellcolor[gray]{0.9}96.21   &    50.28   & \cellcolor[gray]{0.9}92.62 \\
			LatDiff+SG3 & Xception \cite{Xception} &  51.51&   \cellcolor[gray]{0.9}94.23    &   50.14   &   \cellcolor[gray]{0.9}36.27    &    53.64  &   \cellcolor[gray]{0.9}98.68  &    64.77   &   \cellcolor[gray]{0.9}98.34     \\
	\midrule
		\hspace{1.1em}LatDiff & CLIP \cite{clip}  & \multicolumn{1}{c}{62.33} & \multicolumn{1}{c}{\cellcolor[gray]{0.9}74.53} &   52.99    & \cellcolor[gray]{0.9}53.23      &  55.71     &  \cellcolor[gray]{0.9}61.05     & \multicolumn{1}{c}{56.64 } & \multicolumn{1}{c}{\cellcolor[gray]{0.9}64.38 } \\
	LatDiff+SG3 & CLIP \cite{clip} &   65.02    &    \multicolumn{1}{c}{\cellcolor[gray]{0.9}83.65}   &  50.58     &   \cellcolor[gray]{0.9}48.29    &  55.96     &   \cellcolor[gray]{0.9}68.04    &    59.45   &\cellcolor[gray]{0.9}68.76  \\
	\midrule
	
	\hspace{1.1em}LatDiff & MFCLIP &   99.99  & \cellcolor[gray]{0.9}99.99 &   65.08   &     \cellcolor[gray]{0.9}77.07  &  99.98     &  \cellcolor[gray]{0.9}99.98     &    97.92   & \cellcolor[gray]{0.9}99.99 \\
		LatDiff+SG3  & MFCLIP & \multicolumn{1}{c}{\textbf{100}}     & \textbf{ 100 } \cellcolor[gray]{0.9}   &  \textbf{65.10}     &   \textbf{\cellcolor[gray]{0.9}77.84}   &  \textbf{99.99}     & \textbf{\cellcolor[gray]{0.9} 99.99}  &   \textbf{98.32}   &\multicolumn{1}{c}{\textbf{\cellcolor[gray]{0.9}100}} \\
	\bottomrule
\end{tabular}%
	}
\end{table}
\vspace{-1em}
\subsection{ Impacts of the FID distribution in FTG.} To study the correlation between the FID and the criteria for level classifications in FTG, we calculate the FID score of various generators or manipulations at different levels in FTG. To guarantee a fair and balanced assessment, we keep an equivalent distribution of pristine and forgery face images, i.e., a 1:1 ratio. In detail, for level 1, we randomly select 10k real face images and 10k fake ones, with the fake ones evenly distributed across ten deepfake generators (1k images per generator). Similarly, at levels 2 and 3, we choose an equal number of real and fake images to calculate the FID scores for each forgery, respectively. At level 4, we also adopt 10k real images and 10k fake ones to compute the FID score of each generator.

The results across various levels in FTG are displayed in Table~\ref{fid}. We observe that the difference in FID scores across categories under each level becomes increasingly pronounced, as the hierarchy goes deeper. This phenomenon suggests that classification may become easier, as the number of categories increases from 2 (L1) to 4 (L2), 7 (L3), and 11 (L4) \cite{genface}. Moreover, the FID scores exhibit a corresponding rise from 13.097 (L1) to 20.585 (L2), 29.079 (L3), and 30.739 (L4), with the hierarchy level increasing, which facilitates the deepfake detection since there is a growing distinction between authentic and forged images. It is observed that the FID score of the diffusion mode is lower than that of GAN under different forgeries in L3, indicating that diffusion face images achieve superior quality. Therefore, we argue that the diffusion mode is inclined to push detectors to study more discriminative and general forgery artifacts, thus improving generalization capabilities.

 We notice that when the distribution is the most different from the others in GAN data, the FID score is improved.  To study the impact of different distributions on model performance, face images created by LatDiff and StyleGAN3 with the largest distribution (i.e. FID) difference under the EFS forgery are selected to train detectors. In Table~\ref{data}, the AUC of the detector trained using both LatDiff and StyleGAN3 outperforms that of the model trained with LatDiff. This suggests that the performance of detectors tends to increase, when facial images with significant distributional differences are introduced. We argue that the new distribution may increase the diversity of data, allowing the model to learn more robust features, and thereby improving generalization capabilities.
\vspace{-1em}
\subsection{Influences of loss functions}
To verify the contribution of loss functions, we perform ablations on various losses. The ablation result of losses is displayed in Table~\ref{tabloss}. As MFCLIP is guided with merely cross-entropy loss, the AUC is 95.48\% on LatDiff, but an about 1.8\% increase of AUC could be reached via adding the KL loss. We believe that it could enhance the visual forgery representations via language guidance. Meanwhile, due to the introduction of the CMC loss, MFCLIP is grown by 2.6\% AUC on LatDiff, showing the effectiveness of adaptive cross-modal pairs alignment. The integration of the three losses performs the best among these losses, which suggests that the proposed loss could acquire promising results. 
	
\begin{table}[t!]
	\caption{ Effects of pre-trained weights of CLIP. † denotes the image encoder followed by the image adapter or image LoRA. * denotes the text encoder followed by a text adapter or text LoRA. ‡ denotes both image and text encoders, each followed by an adapter or LoRA. We test models on DDPM, LatDiff, DiffFace, and Diffae, after training on CollDiff in GenFace. FFT means full fine-tuning. \label{tabpretrain}}
	\setlength{\tabcolsep}{1.1mm}{
	\begin{tabular}{ccccccccc}
			\toprule
		\multirow{2}[3]{*}{Model} & \multicolumn{2}{c}{DDPM} & \multicolumn{2}{c}{LatDiff} & \multicolumn{2}{c}{DiffFace} & \multicolumn{2}{c}{Diffae} \\\cmidrule{2-9}      & ACC   & AUC   & ACC   & AUC   & ACC   & AUC   & ACC   & AUC \\\midrule
		CLIP & 51.58 & 53.92 & 50.21 & 46.17  & 50.26 & 48.64 & 49.82 & 48.36\\CLIP-FFT   & 62.38 &84.01 &50.02 & 56.39 & 51.91 & 70.28 & 50.51 & 60.62\\
		CLIP-Adapter†   & 70.66 & 78.06 & 48.32 & 45.06 & 59.32 & 70.74 & 55.40 & 62.51 \\ CLIP-Adapter* & 63.56 & 70.38 & 48.06 & 44.30 & 65.16 & 73.04 & 58.19 & 57.49 \\ CLIP-Adapter‡  & 65.10 & 80.27 & 48.79  & 43.45 & 63.48 & 72.32 &64.97 & 67.76 \\ CLIP-LoRA†   & 80.52 & 84.66 & 50.78 & 49.07 & 68.64 & 75.51 & 70.98 & 75.11 \\ 	CLIP-LoRA*   & 81.87 & 84.93 & 46.97 & 44.20 & 68.02 & 75.54 & 67.54 & 74.37 \\  CLIP-LoRA‡  & 80.36 & 84.63 & 48.67  & 45.02 & 68.49 & 75.08 & 68.96 & 73.01  \\MFCLIP & \textbf{100}   & \textbf{100}   & \textbf{99.19} & \textbf{99.97} & \textbf{99.63} & \textbf{99.96}  & \textbf{93.94}& \textbf{99.99} \\\bottomrule
	\end{tabular}%
	}
\end{table}
\vspace{-1em}
\subsection{Impacts of pre-trained weights of CLIP} 

To study the effect of pre-trained weights of CLIP, we fine-tune pre-trained CLIP models using face images generated by CollDiff, and test them on DDPM, LatDiff, DiffFace, and Diffae. Specifically, we adjust all parameters of the pre-trained CLIP model via full fine-tuning. To conduct parameter-efficient fine-tuning, we incorporate residual-style adapters or rank decomposition matrices at the end of CLIP’s image and text encoder, respectively, following CLIP-Adapter \cite{clipadapter} and LoRA \cite{lora} (rank=8, alpha=32). We freeze pre-trained CLIP's image and text encoders, and fine-tune using either the image adapter, image LoRA, text adapter, or text LoRA, accordingly. We also fine-tune CLIP using both the text and image adapters or both the text and image LoRA. In Table~\ref{tabpretrain}, it is noticed that CLIP fine-tuned with LoRA achieves the best performance among various fine-tuned CLIPs. Besides, we can observe that the performance of CLIP with the updated image adapter outperforms that of CLIP with the modified text adapter. We argue that image feature adaptation is more important than text feature adjustment for DFFD, since the domain gap of visual embeddings between the pre-trained dataset and face forgery dataset is bigger than that of text representations. 

Besides, we train the vanilla CLIP from scratch without loading pre-trained weights. Table~\ref{tabpretrain} shows that the performance of fine-tuned CLIPs is higher than that of the CLIP trained from scratch. We believe that fine-tuning enables CLIP to effectively utilize the feature representations learned during pre-training on large-scale datasets. This allows for improved generalization, particularly for tasks involving subtle distinctions, such as FFD. The performance of our MFCLIP method outperforms that of fine-tuned CLIP models, demonstrating that both global image-noise fusion and flexible vision-language matching are helpful for DFFD.
\vspace{-1em}
\section{Visualization}\label{sec6}
{\bfseries\setlength\parindent{0em} Visualization of SPA.}
To demonstrate the effectiveness of our SPA method, we visualize the cosine similarity matrix, when SPA is involved or not. As Fig.~\ref{figspa} (a) displays, relevant negative pairs acquire larger correlation scores and vice versa, due to the addition of SPA, which shows that SPA emphasizes the related negative pairs and suppresses the irrelevant ones to achieve flexible alignment.

{\bfseries\setlength\parindent{0em} Visualization of heatmap.}
To further investigate the effect of MFCLIP, we display the heatmap of different detectors in Fig.~\ref{figspa} (b). Each row shows a forgery face yielded by various generators. The second to fifth columns illustrate heatmaps of four models: (I) CLIP; (II) DIRE; (III) FreqNet, and (IV) MFCLIP. Compared to other detectors, MFCLIP (III) captures more manipulated areas, showing that text-guided image-noise face forgery representation learning can benefit FFD. Specifically, in the last row, the pristine face is added with bangs through the attribute-manipulated model Diffae, and we notice that the MFCLIP model could, to a large extent, identify the forgery area.
\vspace{-1em}
\section{Conclusion}\label{sec7}

We propose a novel MFCLIP method to facilitate the advancement of generalizable DFFD. First, we build the fine-grained text generator to produce the text prompts of each image in GenFace. Second, we observe that the significant discrepancy between the authentic and forgery facial SRM noises extracted from the richest patches, compared to the poorest patches, we design the noise encoder to capture the discriminative and fine-grained noise forgery patterns from the richest patches. Furthermore, we devise the fine-grained language encoder to extract the abundant text embeddings. We also present a novel plug-and-play SPA method to align features of cross-modal sample pairs, adaptively, which could be integrated into any vision-language-based model like CLIP with only a slight growth in computational costs.

{\bfseries\setlength\parindent{0em} Limitations.} Although our model has explored fine-grained multi-modal forgery traces, we need to improve generalization across various generators and reduce computational complexity. In the future, to solve the first restriction, fine-tuning the detector with face forgery data generated by advanced methods, such as autoregressive and flow-based models, could improve the generalizability and reliability. To overcome the second limitation, we are expected to devote ourselves to model architecture optimization such as lightweight model design and module replacement, for faster efficiency. For the former, reducing the number of transformer layers, hidden dimensions, or attention heads could be realized to decrease the complexity of the proposed model. In addition, a simpler and faster student model could be trained to mimic the behaviour of the proposed MFCLIP model via knowledge distillation. For the latter, linear attention or other efficient attention variants (e.g., vision mamba \cite{mamba} or spike-driven self-attention \cite{Spike}) could be employed to reduce the computational complexity of the transformer in MFCLIP.
\vspace{-1em}
\bibliographystyle{IEEEtran}
\bibliography{mfclip}
\vspace{-4em}
\begin{IEEEbiography}[{\includegraphics[width=1in,clip,keepaspectratio]{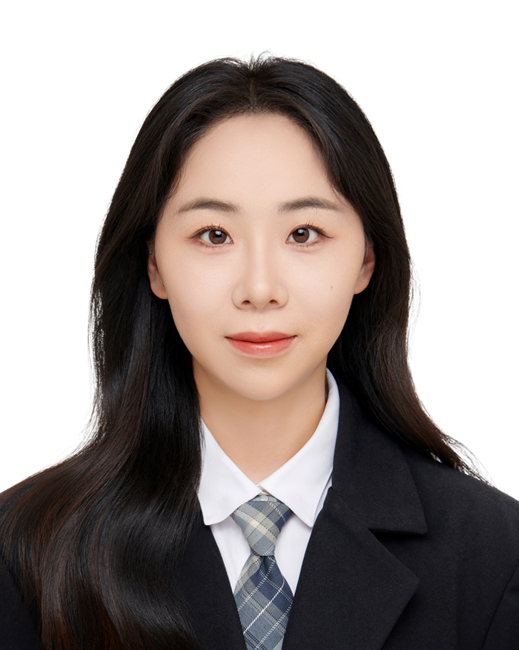}}]{Yaning Zhang} received the double B.S. degree in Internet of Things Engineering and English in 2020, and the M.S. degree in Computer Applied Technology in 2023, both from Qilu University of Technology (Shandong Academy of Sciences), Jinan, China, where she is currently pursuing the Ph.D. degree. Her research interests include computer vision, artificial intelligence, multimedia forensics, and face forgery detection.
\end{IEEEbiography}
\vspace{-2em}
\begin{IEEEbiography}[{\includegraphics[width=1in,clip,keepaspectratio]{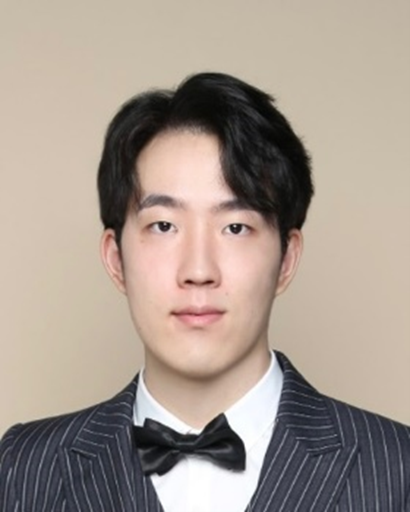}}]{Tianyi Wang} (Member, IEEE) received the double major B.S. degree in Computer Science and Applied and Computational Mathematical Sciences from the University of Washington, Seattle, USA, in 2018. After that, he received the Ph.D. degree in Computer Science, under the supervision of Dr. Kam Pui Chow, from the University of Hong Kong, Hong Kong, in 2023. He is currently a Research Fellow at School of Computing, National University of Singapore, Singapore. His major research interests include multimedia forensics, face forgery detection, misinformation detection, generative artificial intelligence, and computer vision.
\end{IEEEbiography}
\vspace{-3em}
\begin{IEEEbiography}[{\includegraphics[width=1in,clip,keepaspectratio]{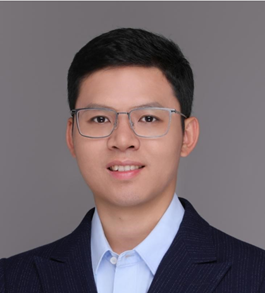}}]{Zitong Yu } (Senior Member, IEEE) received the Ph.D. degree in Computer Science and Engineering from the University of Oulu, Finland, in 2022. Currently, he is an Assistant Professor at Great Bay University, China. He was a Postdoctoral researcher at ROSE Lab, Nanyang Technological University. He was a visiting scholar at TVG, University of Oxford. His research interests include human-centric computer vision and biometric security. He was area chairs of ICME 2023, BMVC 2024, and IJCB 2024. He was a recipient of IAPR Best Student Paper Award, IEEE Finland Section Best Student Conference Paper Award 2020, Best Paper Candidate of ICME 2024, second prize of the IEEE Finland Jt. Chapter SP/CAS Best Paper Award (2022), and World's Top 2\% Scientists 2023 by Stanford University.
\end{IEEEbiography}
\vspace{-2em}
\begin{IEEEbiography}[{\includegraphics[width=1in,clip,keepaspectratio]{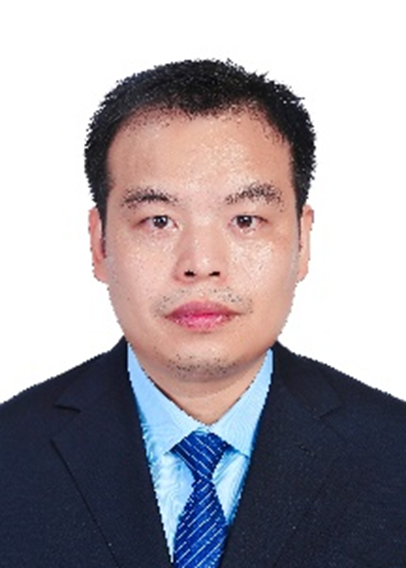}}]{Zan Gao}  (Senior Member, IEEE) received his Ph.D degree from Beijing University of Posts and Telecommunications in 2011. He is currently a full Professor with the Shandong Artificial Intelligence Institute, Qilu University of Technology (Shandong Academy of Sciences). From Sep. 2009 to Sep. 2010, he worded in the School of Computer Science, Carnegie Mellon University, USA. From July 2016 to Jan 2017, he worked in the School of Computing of National University of Singapore. His research interests include artificial intelligence, multimedia analysis and retrieval, and machine learning. He has authored over 100 scientific papers in international conferences and journals including TPAMI, TIP, TNNLS, TMM, TCYBE, CVPR, ACM MM, WWW, SIGIR and AAAI.
\end{IEEEbiography}

\vspace{-2em}
\begin{IEEEbiography}[{\includegraphics[width=1in,clip,keepaspectratio]{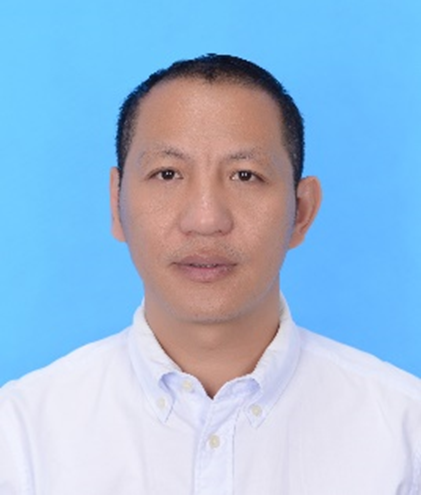}}]{Linlin Shen} (Senior Member, IEEE) is currently a Pengcheng Scholar Distinguished Professor at School of Computer Science and Software Engineering, Shenzhen University, Shenzhen, China. He is also an Honorary professor at School of Computer Science, University of Nottingham, UK. He serves as the Deputy director of National Engineering Lab of Big Data Computing Technology, Director of Computer Vision Institute, AI Research Center for Medical Image Analysis and Diagnosis, and China-UK joint research lab for visual information processing. He also serves as the Co-Editor-in-Chief of the IET journal of Cognitive Computation and Systems and Associate Editor of Expert Systems With Applications. His research interests include deep learning, facial recognition, analysis/synthesis and medical image processing. Prof. Shen is listed as the “Most Cited Chinese Researchers” by Elsevier, and listed in a ranking of the “Top 2\% Scientists in the World” by Stanford University. He received the “Best Paper Runner-up Award” from the journal of IEEE Transactions on Affective Computing, and “Most Cited Paper Award” from the journal of Image and Vision Computing. His cell classification algorithms were the winners of the International Contest on Pattern Recognition Techniques for Indirect Immunofluorescence Images held by ICIP and ICPR. His team has also been the runner-up and second runner-up of a number of competitions for object detection in remote sensing images, nucleus detection in histopathology images and facial expression recognition.
\end{IEEEbiography}
\vspace{-2em}
\begin{IEEEbiography}[{\includegraphics[width=1in,clip,keepaspectratio]{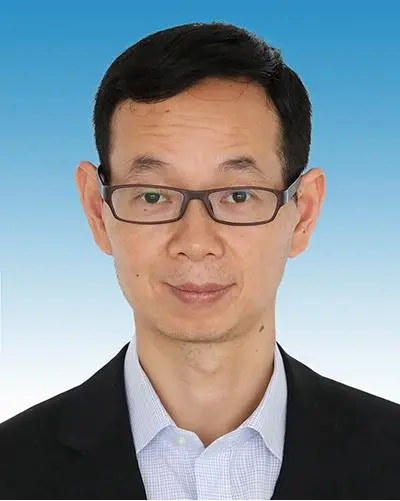}}]{Shengyong Chen} received the Ph.D. degree in computer vision from the City University of Hong Kong, Hong Kong, in 2003. He was with the University of Hamburg from 2006 to 2007. He is currently a Professor at the Tianjin University of Technology and also with the Zhejiang University of Technology, China. He received the Fellowship from the Alexander von Humboldt Foundation of Germany. He has authored over 100 scientific papers in international journals. His research interests include computer vision, robotics, and image analysis. He is a fellow of the IET and a Senior Member of the CCF. He received the National Outstanding Youth Foundation Award of China in 2013.
\end{IEEEbiography}
\end{document}